\documentclass[a4paper, 12pt, oneside]{article}

\pdfoutput=1

\usepackage{fancyhdr}
\usepackage{setspace}
\usepackage[top=3.5cm, bottom=3cm, left=2cm, right=2cm]{geometry}
\usepackage{hyperref}
\hypersetup{
	unicode=true,
	pdftitle={Machine Reasoning Explainability},
	pdfauthor={Ericsson Research},
	pdfsubject={Explainable AI and Machine Reasoning},
	pdfkeywords={Explainable AI, Machine Reasoning},
	colorlinks=true,
	linkcolor=black,
	citecolor=black,
	urlcolor=black,
	filecolor=black,
	pdfstartview=FitH,
	pdfborder={0 0 0}
}

\usepackage[utf8x]{inputenc}
\usepackage{times}  
\usepackage{helvet}  
\usepackage{courier}  

\usepackage{hyperref}
\usepackage{url}  

\usepackage{graphicx}

\usepackage{tikz}

\usepackage{amssymb}
\usepackage{amsmath}
\usepackage{mathtools}
\usepackage{float}
\usepackage{multicol}
\usepackage[inline, shortlabels]{enumitem}
\setlist{nolistsep, leftmargin=*}
\SetEnumerateShortLabel{q}{Q\arabic*}

\usepackage[square,sort,numbers,longnamesfirst]{natbib}

\usepackage{macros} 

\title{Machine Reasoning Explainability}
\author{Kristijonas \v{C}yras\thanks{Corresponding author. Email: \href{kristijonas.cyras@ericsson.com}{\tt{kristijonas.cyras@ericsson.com}}, ORCiD: \href{https://orcid.org/0000-0002-4353-8121}{0000-0002-4353-8121}}, Ramamurthy Badrinath, Swarup Kumar Mohalik, \\  Anusha Mujumdar, Alexandros Nikou, Alessandro Previti, \\
	Vaishnavi Sundararajan, Aneta Vulgarakis Feljan \\ \textbf{Ericsson Research}}
\date{\today}

\begin{document}

\pagestyle{fancy}
\setlist{nolistsep}

\maketitle

\begin{abstract}
As a field of AI, Machine Reasoning (MR) uses largely symbolic means to formalize and emulate abstract reasoning. 
Studies in early MR have notably started inquiries into Explainable AI (XAI) -- arguably one of the biggest concerns today for the AI community. 
Work on explainable MR as well as on MR approaches to explainability in other areas of AI has continued ever since. 
It is especially potent in modern MR branches such as argumentation, constraint and logic programming, and planning. 
We hereby aim to provide a selective overview of MR explainability techniques and studies in hopes that insights from this long track of research will complement well the current XAI landscape. 
\end{abstract}


\pagebreak
\tableofcontents

\section{Introduction}
\label{sec:Introduction}

\emph{Machine Reasoning} (MR) is a field of AI that complements the field of Machine Learning (ML) by aiming to computationally mimic abstract thinking. This is done by way of uniting known (yet possibly incomplete) information with background knowledge and making inferences regarding unknown or uncertain information. MR has outgrown Knowledge Representation and Reasoning (KR, see e.g.~\cite{Brachman:Levesque:2004}) and now encompasses various symbolic and hybrid AI approaches to automated reasoning. 
Central to MR are two components: 
a knowledge base (see e.g.~\cite{Davis:2017})
or a model of the problem (see e.g.~\cite{Geffner:2018})
, which formally represents knowledge and relationships among problem components in symbolic, machine-processable form; and a general-purpose inference engine or solving mechanism, which allows to manipulate those symbols and perform semantic reasoning.\footnote{See~\cite{Bottou:2014} for an alternative view of MR stemming from a sub-symbolic/connectionist perspective.}

The field of Explainable AI (XAI, see e.g.~\cite{Neches:Swartout:Moore:1985,Swartout:Paris:Moore:1991,Core.et.al:2006,Langley.et.al:2017,Biran:Cotton:2017,Adadi:Berrada:2018,Preece:2018,Mittelstadt:Russell:Wachter:2019,Rosenfeld:Richardson:2019,Arrieta.et.al:2020,Mohseni:Zarei:Ragan:2020}) encompasses endeavors to make AI systems intelligible to their users, be they humans or machines. 
XAI comprises research in AI as well as interdisciplinary research at the intersections of AI and subjects ranging from Human-Computer Interaction (HCI)~\cite{Mohseni:Zarei:Ragan:2020} to social sciences~\cite{Miller:2019,Byrne:2019}. 

According to \citeauthor{Hansen:Rieger:2019}~\cite{Hansen:Rieger:2019}, explainability was one of the main distinctions between the $1^{\rm{st}}$ wave (dominated by KR and rule-based systems) and the $2^{\rm{nd}}$ wave (expert systems and statistical learning) of AI, with expert systems addressing the problems of explainability and ML approaches treated as black boxes. 
With the ongoing $3^{\rm{rd}}$ wave of AI, ML explainability has received a great surge of interest~\cite{Core.et.al:2006,Arrieta.et.al:2020,Mohseni:Zarei:Ragan:2020}. 
By contrast, it seems that a revived interest in MR explainability is only just picking up pace (e.g.\ ECAI 2020 Spotlight tutorial on Argumentative Explanations in AI\footnote{\href{https://www.doc.ic.ac.uk/~afr114/ecaitutorial/}{https://www.doc.ic.ac.uk/$\sim$afr114/ecaitutorial/}} and KR 2020 Workshop on Explainable Logic-Based Knowledge Representation\footnote{\href{https://lat.inf.tu-dresden.de/XLoKR20/}{https://lat.inf.tu-dresden.de/XLoKR20/}}).
However, explainability in MR dates over four decades~\cite{Neches:Swartout:Moore:1985,Swartout:Paris:Moore:1991,Johnson:Johnson:1993,Moulin.et.al:2002,Hansen:Rieger:2019} and can be roughly outlined thus. 

The $1^{\rm{st}}$ generation expert systems provide only so-called (\emph{reasoning}) \emph{trace explanations}, showing inference rules that led to a decision. A major problem with trace explanations is the lack of ``information with respect to the system’s general goals and resolution strategy"\cite[p.~174]{Moulin.et.al:2002}. 
The $2^{\rm{nd}}$ generation expert systems instead provide so-called \emph{strategic explanations}, 
``displaying system’s control behavior and problem-solving strategy.''\cite[p.~95]{Yetim:2005}
Going further, so-called \emph{deep explanations} separating the domain model from the structural knowledge have been sought, where ``the system has to try to figure out what the user knows or doesn’t know, and try to answer the question taking that into account."\cite[p.~73]{Walton:2004} 
Progress in MR explainability notwithstanding, it has been argued~\cite{Moulin.et.al:2002,Lim:Dey:2010,Preece:2018} that to date, explainability in MR particularly and perhaps in AI at large is still insufficient in aspects such as justification (``describing the rationale behind each inferential step taken by the system''~\cite[p.~95]{Yetim:2005}), criticism, and cooperation. 
These aspects, among others, are of concern in the modern MR explainability scene (this millennium), whereby novel approaches to explainability in various branches of MR have been making appearances. 

Explainability is a highly desired aspect of autonomous agents and multi-agent systems (AAMAS)~\cite{Langley.et.al:2017,Kraus.et.al:2020,Murukannaiah.et.al:2020}. 
There are a few area-specific reviews of explainability in AAMAS-related areas: 
for instance~\cite{Anjomshoae.et.al:2019} on human-robot interaction;  \cite{Nunes:Jannach:2017} on expert and recommender systems; 
\cite{Rosenfeld:Richardson:2019} on explaining ML agents; 
\cite{Chakraborti.et.al:2020} on planning.
However, explainability in multi-agent systems (MAS) is still under-explored~\cite{Langley.et.al:2017,Kraus.et.al:2020,Murukannaiah.et.al:2020}.
In AI-equipped MAS (sometimes also called Distributed AI~\cite{Lecue:2020}), explainability concerns interactions among multiple intelligent agents, be they human or AI, to agree on and explain individual actions/decisions. 
Such interactions are often seen as a crucial driver for the real-world deployment of trustworthy modern AI systems. 
We will treat the following as a running example in Section~\ref{sec:Explainability} to illustrate the kinds of explanations that we encounter in the XAI literature, including but not limited to MR approaches. 

\begin{example}
\label{ex:5G}
\label{ex:main}
In modern software-defined telecommunication networks, network slicing is a means of running multiple logical networks on top of a shared physical network infrastructure~\cite{Bega.et.al:2020}. 
Each logical network, i.e.\ a slice, is designed to serve a defined business purpose and comprises all the required network resources, configured and connected end-to-end. 
For instance, in a 5G network, a particular slice can be designated for high-definition video streaming. 
Such network slices are then to be managed---in the future, using autonomous AI-based agents---to consistently provide the designated services. 
Service level agreements stipulate high-level intents that must be met, such as adequate quality of service, that translate into quantifiable performance indicators. 
An example of an intent is that end-to-end latency (from the application server to the end-user) should never exceed 25ms. 
Such intents induce lower level goals that AI-based agents managing the slice need to achieve. 

We consider the following agents to be involved in automatically managing the slice. 
When proactively monitoring adherence to intents, predictions of e.g.\ network latency are employed. 
So first, prediction of latency in the near future (say 20min interval) is done by an ML-based \emph{predictor} agent based on previous network activity patterns (see e.g.~\cite{Rusek.et.al:2020}). 
Given a prediction of latency violation, the goal is to avoid it. 
To this end, a rule-based \emph{root cause analysis} (RCA) agent needs to determine the most likely cause(s) of the latency violation which may, for instance, be a congested router port. 
Given a root cause, a \emph{constraint solver} agent aims to find a solution to a network reconfiguration problem, say a path through a different data centre, that satisfies the slice requirements, including latency. 
Finally, a \emph{planner} agent provides a procedural knowledge-based plan for execution of the reconfiguration (i.e.\ how to optimally relocate network resources). 

In all of the above phases, explainability of the AI-based agents is desirable. 
First and foremost, one may want to know which features contributed the most to the predicted latency violation. 
These may point to the performance measurement counter readings that via domain expert-defined rules lead to inferring the root cause. 
Explaining RCA by indicating the facts and rules that are necessary and sufficient to establish the root cause contributes to the overall explainability of handling intents. 
Orthogonally, the constraint solver may be unable to find a solution within the initial soft constraints, such as limited number of hops, whence the network reconfiguration problem unsolvability could be explained by indicating a set of mutually unsatisfiable constraints and suggesting a relaxation, such as increasing the number of hops. 
When some reconfiguration solution is found and the planner yields a plan for implementation, its goodness as well as various alternative actions and contrastive states may be considered for explainability purposes. 
Last but not least, the overall decision process needs to be explainable too, by for instance exhibiting the key considerations and weighing arguments for and against the best outcomes in all of the phases of prediction, RCA, solving and planning.
\end{example}

MR has a particularly strong presence in AAMAS and a number of recent works specifically tackle explainability: 
see, for instance, \cite{Sklar:Azhar:2018,Madumal.et.al:2019,Karamlou:Cyras:Toni:2019-demo,Raymond:Gunes:Prorok:2020} for human-agent interaction, 
\cite{Fan:2018,Kulkarni.et.al:2019,Torta.et.al:2019,Eifler.et.al:2020-IJCAI} for autonomous agent planning and~\cite{Cyras.et.al:2020-AAMAS} for scheduling,  \cite{Zeng.et.al:2018-AAMAS,Zhong.et.al:2019,Boixel:Endriss:2020} for multi-agent decision making, 
\cite{Cocarascu:Rago:Toni:2019} for argumentative dialogical agents. 
However, overviews of MR approaches to explainability, in AAMAS and XAI generally, are lacking. 
In this report we aim to fill the gap. 
To this end, we provide a conceptual overview of MR explainability techniques, including but not limited to those applicable to AAMAS. 
We look into the kinds of questions explanations provided by MR approaches answer and loosely categorize explanations, arguing that they are by and large \emph{attributive} or \emph{contrastive}, but not yet (often) \emph{actionable}. 
We suggest that explainability in AAMAS and AI at large would greatly benefit from MR explainability insights and techniques.\footnote{Echoing Randy Goebel's invited talk on XAI at ECAI 2020.}

\subsection{Contributions}
\label{subsec:Contributions}

Our contributions in this report are as follows. 
\begin{enumerate}
    \item Building on conceptual works on XAI as well as XAI overviews, we propose a loose categorization of MR explainability approaches in terms of broad families of explanations, namely \emph{attributive}, \emph{contrastive} and \emph{actionable}. 
    \item We provide an overview of some prominent approaches to MR explainability, differentiating by branches of MR and indicating the families that such explanations roughly belong to. 
    \item We indicate what kinds of questions the explanations aim to answer. 
\end{enumerate}

This is not a systematic review, but rather an overview of conceptual techniques that MR brings to XAI. 
We do not claim to be exhaustive or even completely representative of the various MR approaches to explainability, let alone to characterize what counts as a branch of MR. 
Rather, we hope to enable the reader to see a bigger picture of XAI, focusing specifically on what we believe amounts to MR explainability.

\subsection{Motivations}
\label{subsec:Motivation}

The following summarizes our motivations and assumptions in this work.

\begin{enumerate}
    \item We appreciate that MR is not yet a commonplace AI term, unlike e.g.\ ML or KR. We recognize that MR has evolved from KR and comprises other, mostly symbolic, forms of reasoning in AI. However, we here do not attempt to characterize MR, let alone cover all of its branches. Rather, we focus on the MR branches that most prominently exhibit approaches to explainability. Our overview of MR explainability is therefore bounded in scope. Nonetheless, it is dictated by our (invariably limited) professional understanding of the most relevant (e.g.\ historically important, well established and widely applicable, or trending) MR contributions to XAI. 

    \item Accordingly, we acknowledge that explainability in AI has been studied for a while and has in time evolved in terms of 
    \begin{enumerate*}
        \item areas of AI and
        \item desiderata.
    \end{enumerate*}
    Yet, we maintain that the foundational contributions and the lessons learnt (as well as forgotten) from the research on explainability in KR and expert systems are still very much relevant to both MR explainability in particular, and XAI at large. Specifically: 
    \begin{enumerate}
        \item Adaptations of long established MR explainability techniques (\nameref{subsec:Inference} and~\nameref{subsec:LP}; see Sections~\ref{subsec:Inference} and~\ref{subsec:LP} respectively) 
        have found their ways into newer MR areas (for instance, \nameref{subsubsec:pinpointing} in Description Logics,  \nameref{subsec:Planning}; see Sections~\ref{subsubsec:pinpointing} and~\ref{subsec:Planning} respectively). 
        
        \item Relatively newer MR branches, such as~\nameref{subsubsec:ASP} (Section~\ref{subsubsec:ASP}) and~\nameref{subsec:Argumentation} (Section~\ref{subsec:Argumentation}), necessitate and inform newer forms of explainability. 
        In particular, previously established techniques of mostly logical inference attribution do not suffice anymore~\cite{Moulin.et.al:2002,Walton:2004,Kulesza.et.al:2015,Freuder:2017}. 
        On the one hand, some modern MR approaches, such as ASP and~\nameref{subsec:CP} (Section \ref{subsubsec:CP}), currently provide techniques that can effectively be considered interpretable but whose workings are nevertheless difficult to explain (see Section~\ref{subsec:Terminology} for some remarks on terminology).
        On the other hand, XAI desiderata now include aiming for dialogical/conversational explanations, interactivity, actionability as well as causality. 
        These rediscovered concerns are being addressed by modern MR approaches to explainability. 
        Thus, various branches of MR are concerned with explainability anew. 
    \end{enumerate}
	
    \item We maintain that modern MR systems can hardly be called explainable \emph{just by virtue of being symbolic}\footnote{Where symbolic entities carry intrinsic semantic meaning and are perhaps more readily interpretable and intelligible than the algebraic-symbolic entities in sub-symbolic/connectionist AI.}, 
    in contrast to early expert and intelligent systems often being referred to thus~\cite{Adadi:Berrada:2018}. 
    However, we speculate and contend that MR presumably offers more immediate intelligibility than e.g.\ ML. 
    Perhaps this is what lends MR explainability approaches to be applied to \emph{both} MR itself and other areas of research, 
    such as ML~\cite{Shih:Choi:Darwiche:2018,Ignatiev.et.al:2019,Cocarascu.et.al:2020,Albini.et.al:2020,Bertossi:2020}, 
    decision support and recommender systems~\cite{Nunes:Jannach:2017,Fox:2017,Rago:Cocarascu:Toni:2018,Chapman.et.al:2019}, 
    planning~\cite{Fox:Long:Magazzeni:2017,Fan:2018,Cashmore.et.al:2019,Eifler.et.al:2020-AAAI,Chakraborti.et.al:2020}), 
    scheduling~\cite{Cyras.et.al:2019-AAAI}, 
    legal informatics~\cite{Bex:2015,Cyras.et.al:2019-ESWA,Collenette:Atkinson:Bench-Capon:2020}, 
    scientific debates~\cite{Seselja:Strasser:2013}. 
    We speculate that it is also potentially the theoretical guarantees often ensured by MR methods that make MR explainability appealing, cf.\ e.g.~\cite{Slack.et.al:2020,Ignatiev:2020-IJCAI,Dimanov.et.al:2020}. 
    
    \item Most recent overviews of general XAI appear to focus mostly on ML and somewhat ignore MR, e.g.~\cite{Adadi:Berrada:2018,Guidotti.et.al:2019,Rosenfeld:Richardson:2019,Mittelstadt:Russell:Wachter:2019,Arrieta.et.al:2020,Mohseni:Zarei:Ragan:2020}, apart from potentially briefly discussing early expert systems. 
    (See however~\cite{Nunes:Jannach:2017} for a systematic review of explanations in decision support and recommender systems, where MR constitutes majority of the referenced approaches; 
    reviews of explainability in e.g.\ ASP~\cite{Fandinno:Schulz:2019} and \nameref{subsec:Planning}~\cite{Chakraborti.et.al:2020} are also welcome examples of not-so-common area-specific MR overviews.) 
    We feel that having a broader view of the XAI agenda is crucial in general and that an overview of MR explainability is due for at least the following reasons.
    \begin{enumerate}
        \item Explainable MR constitutes a rich body of research whose works span many branches of MR with long histories. A consolidated, even if limited, overview will provide guidance to exploring and building on this research. 
        \item MR explainability techniques are also used for e.g. explainable ML and an overview will help AI researchers to see a bigger picture of XAI as well as to promote innovation and collaboration. 
        \item XAI has arguably started with explainable MR and some of the same conceptual MR explainability techniques can be (and are being) applied to achieve current XAI goals. 
        An MR explainability overview may allow researchers to rediscover known problems and solutions (potentially saving from reinventing things) and to give credit where it is due. 
        (See e.g.~\cite{Abdul.et.al:2018} for a discussion on findings that the current trend of explainability in ML is not well connected to earlier XAI research.)
    \end{enumerate}
\end{enumerate}

\section{Explainability}
\label{sec:Explainability}

We here briefly discuss the main purposes of explanations in AI systems and present a categorization of explanations. 
We propose that, intuitively, the main purpose of explanations in XAI is to enable the user of an AI system to not only understand (to a certain extent) the system, but also to do something with the explanation. 
We suggest this entails that the explanations answer some sort of questions about the AI system dealing with the problem at hand. 
The kinds of answers provided by the explanations will help us to loosely categorize them, enabled by approaches to explainability in MR.

\subsection{Purpose of Explanations}
\label{subsec:Purpose}

In general, the purpose of explainability of an AI system can be seen to be two-fold, to quote from~\cite[p.~160]{Johnson:Johnson:1993}:
\begin{quote}
The system either provides knowledge and explanations necessary for the user to carry out his or her task, or alternatively, the system carries out some action and then explains the need and reason for the action the system itself has taken to the user.
\end{quote}
Borrowing from~\cite[p.~85, emphasis ours]{Arrieta.et.al:2020}, we state what an explainable AI system entails:
\begin{quote} 
Given an audience, an explainable Artificial Intelligence is one that \emph{produces details or reasons} to make its functioning clear or easy to understand.
\end{quote}
We thus stipulate that an explainable AI system has to first and foremost \textbf{produce details and reasons underlying its functioning and outputs thereof} -- call this an \textbf{explanation}. 
It is then upon the explainee (i.e.\ the recipient or user of the explanation, also called the audience) to take in the explanation. Thus, we are interested in the purpose of an explanation (see e.g.~\cite{Nunes:Jannach:2017,Ribera:Lapedriza:2019,Srinivasan:Chander:2020,Mohseni:Zarei:Ragan:2020}) from the explainee point-of-view: 
\begin{quote}
    ``What will I (i.e.\ the agent, human or artificial) do with the explanation?" 
\end{quote}

We do not attempt a representative list of the purposes of explanations, but nevertheless give some examples. For instance, an expert in the domain in which an AI system is applied may want to do any one of the following:
\begin{enumerate}[a)]
    \item Understand how the system functions in general, in mundane situations, whence they expect explanations to pertain to certain known aspects of the domain;
    \item Learn how the system outputs aspects of the domain that are unexpected to the expert; 
    \item Confirm and compare the system’s behavior in both expected and unexpected situations;
    \item Act on the produced knowledge, whence they expect guidance towards desirable results.
\end{enumerate}

Different purposes of explanations are well reflected by what are called XAI design goals in \cite{Mohseni:Zarei:Ragan:2020}, that amount to “identifying the purpose for explanation and choosing what to explain for the targeted end-user and dedicated application.” 
See also e.g.\ \cite{Nunes:Jannach:2017,Srinivasan:Chander:2020} for purposes of, particularly, human-centric explanations in AI. 
Note, however, that we do not limit ourselves to human users of AI systems. 
Instead, we allow AI (or otherwise intelligent machine) agents themselves to require explanations from AI systems. Considering the progress in developing intelligent machines and autonomous systems, it seems natural to foresee situations where an AI agent probes another for e.g.\ a justification of the latter’s (intended) actions in order to achieve or negotiate common goals. 

We do acknowledge that the intended purpose is part of the explanation’s context, among other factors such as the explainer, the explainee and the communication medium \cite{Srinivasan:Chander:2020}. The context, specifically the explainee, may inform or correlate with the purpose of explanations, especially in cases of human audiences \cite{Tomsett.et.al:2018,Arrieta.et.al:2020,Zhou:Danks:2020,Mohseni:Zarei:Ragan:2020}. But that need not generally be the case: if the explainees are people, then it is “people’s goals or intended uses of the technology, rather than their social role” \cite{Zhou:Danks:2020} that shape the explainability desiderata. 
Note that there can be both AI-to-human and AI-to-AI explainability, but we consider the purpose of explanations, rather than the nature of the explainee, to be primary.

Finally, we recognize that various usability desiderata~\cite{Sokol:Flach:2020}, including actionability (see e.g.~\cite{Kulesza.et.al:2015,Cyras.et.al:2020-AAMAS,Sokol:Flach:2020}), are key to the purpose of explainability. We thus maintain that irrespective of the purpose of explanations, for the explainee to consume and act upon, i.e.\ “to do something with” an explanation, \emph{the explanation has to answer some question(s) about the AI system and its functioning}. 
We discuss next how explanations in MR (and potentially in AI at large) can be categorized according to the questions they aim to answer.


\subsection{Categorization for Explanations}
\label{subsec:Categorization}

The driving factors of our categorization are the questions that explanations specifically in various MR approaches aim to answer. 
Intuitively, answering \emph{what}-, \emph{why}-, \emph{when}-, \emph{how}-types of questions about an AI system’s functioning falls under the purview of explainability~\cite{Neches:Swartout:Moore:1985,Lim:Dey:2010,Fox:Long:Magazzeni:2017,Mohseni:Zarei:Ragan:2020}. 
Operationally, consider an AI system which, on some input information, produces a certain outcome. If the outcome is ``expected'', an observer could ask for what properties of the system and the input led to this outcome, or why this particular outcome was chosen over some others. 
If, however, the outcome is not expected, an observer might want to know if one could have obtained the expected outcome with some specific different input information, or how to change the system, so as to obtain the expected outcome with the same input information.

Thus, informally, we would like to semantically differentiate explanations into three different classes based on the questions they seek to answer -- 
explanations for \textbf{what} and \textbf{why} questions (What made/Why did the system reach this outcome?), 
explanations for \textbf{why not} and \textbf{what if} (Why did the system not reach a different outcome? What if different information were used?), 
and explanations for \textbf{how} (How can I modify the system to obtain a more desirable outcome with the existing information?). 
Such questions need not be limited to MR, but may well apply to e.g.\ ML too. 
In the context of our running example (Example~\ref{ex:5G}), we may seek explanations as to what reasons made the ML-based agent to predict, or a rule-based RCA agent to determine the cause of, the latency violation. 
We may also want to know how different the constraint solver's proposed network reconfiguration would be if some constraint (such as number of network hops) were relaxed; 
or why the planner proposes to first reallocate a date center rather than an edge router. 
Ideally, we would like to get explanations as to how to improve the outcomes or the AI agents themselves. 

With a view to making this informal idea concrete, and building on other works that aim at classifications/categorizations of explanations, see~\cite{Johnson:Johnson:1993,Lacave:Diez:2002,Moulin.et.al:2002,Gonul:Onkal:Lawrence:2006,Lim:Dey:2010,Kulesza.et.al:2015,Preece:2018,Rosenfeld:Richardson:2019,Miller:2019,Hansen:Rieger:2019,Mueller.et.al:2019,Wang.et.al:2019,Sokol:Flach:2020,Mohseni:Zarei:Ragan:2020}, we distinguish three families/types of explanations: \textbf{attributive} (e.g.\ logical inference attribution, feature importance association), \textbf{contrastive} (e.g.\ reasons con as well as pro, counterfactuals), \textbf{actionable} (e.g.\ guidelines towards a desired outcome, realizable actions). 
To characterize these families of explanations, we abstractly assume at hand
\begin{quote}
    an AI system \AI\ that given information $i$ yields outcome $o$.\footnote{Note that this is not a formal definition or description. 
    We believe that due to diversity of MR, a formalization of such terms would be incomplete, controversial and possibly unhelpful, if not impractical or outright impossible. Instead, we use natural language descriptions to practically convey our ideas and supply intuition.}
\end{quote}

In MR, for instance, the system can be instantiated by a knowledge base $\KB$ consisting of background knowledge, that given information (e.g.\ a query) $i$ entails (for some formally defined notion $\models$ of entailment) the inference $o$: 
in symbols, $\KB \cup \{ i \} \models o$. 
Examples include: forms of \nameref{subsec:LP} (Section~\ref{subsec:LP}), with $\KB$ a logic program, $i$ a fact or a query and $o$ some inference; 
constraint propagation in \nameref{subsec:CP} (Section~\ref{subsubsec:CP}), with $\KB$ a set of constraints, $i$ a constraint to be propagated and $o$ the propagation result. 
The given information $i$ could well be implicit, as for example in satisfiability problems (Section~\ref{subsubsec:SAT}), with $i$ a (partial) assignment to the formula instantiating the knowledge of the AI system. 
The outcome $o$ could also be implicit, as for example in \nameref{subsubsec:pinpointing} (Section~\ref{subsubsec:pinpointing}),  representing entailment or non-entailment of a given query $Q$ from a knowledge base $\KB$. 
An AI system can also be based on \nameref{subsec:Argumentation} (Section~\ref{subsec:Argumentation}) and instantiated by an argument graph $\graph$, yielding the outcome as to the acceptability of a particular argument $a$ in $\graph$. 
Still further, an autonomous planning system, for instance, can be instantiated by a model $\M$ of the planning problem, yielding a plan $\pi$ given the initial state, goals and plan properties of interest.
In some forms of ML, the role of \AI\ can be taken by a model or function $f : X \to Y$ (trained on some set $X' \subseteq X$ of data), that given instance $i \in X$ assigns to it a label $o \in Y$: in symbols, $f(i)=o$. 

The above are but examples of high-level descriptions of AI methodologies. 
They are nevertheless helpful to find intuitions behind the explanation categories suggested as follows.

\subsubsection{Attributive Explanations}
\label{subsubsec:attributive}

\emph{Attributive explanations} (see e.g.~\cite{Moulin.et.al:2002,Lundberg:Lee:2017,Miller:2019}for conceptual overviews) rely on the notions such as trace, justification, attribution and association, widely used in MR and ML literature alike. At a high level, they aim to answer the following type of question: 
\begin{enumerate}[q, series=questions]
    \item Given a representation of the problem and given a query, e.g.\ a decision taken or an observation, what are the reasons underlying the inference concerning the query?
    \label{what}
\end{enumerate}

Questions of type~\ref{what} pertain to inputs to the system (see e.g.~\cite[p.~14]{Lim:Dey:2010}), definitions and meaning of internal representations (see e.g.~\cite[p.~387]{Neches:Swartout:Moore:1985}, \cite[p.~110]{Lacave:Diez:2002},  \cite[p.~164]{Bussone.et.al:2015}) and attribution of those to the outputs (see e.g.~\cite{Mohseni:Zarei:Ragan:2020}).
Such questions aim at soliciting ``insight about the information that is utilized and the rules, processes or steps that are used by the system to reach the recommendation or the outcome it has generated for a particular case."\cite[p.1483]{Gonul:Onkal:Lawrence:2006} 
Answers to~\ref{what} type of questions thus  ``inform users of the current (or previous) system state" and how the system ``derived its output value from the current (or previous) input values"\cite[p.~14]{Lim:Dey:2010}. 

Using the notions above, attributive explanations answer the following question: 
\begin{quote}
    What are the details and reasons for system \AI\ yielding outcome $o$, given information $i$?
\end{quote}
Attributive explanations thus justify $o$ by attributing to/associating with $o$ (parts of) \AI\ and $i$. 

\begin{example}
\label{ex:attributive}
Consider the root cause analysis (RCA) problem mentioned in Example~\ref{ex:5G}: 
given a prediction (or an observation) that the network latency requirement will be violated due to e.g.~increased packet drop rate, 
determine which parts of the network likely cause the increased packet drop rate. 
Let us assume that the RCA agents implements a rule-based system (compiled from network manuals and expert documentation) $\KB$ and the above observation about packet drop rate is appropriately formalised as a sentence (or set of facts) $i$. 
Suppose that using e.g.\ forward rule chaining the RCA agents finds that some port congestion is a consequence of the knowledge that the agent possesses together with the indicated packet drop rate: 
$\KB \cup \{ i \} \vdash o$, where $o$ represents a congested port. 
An attributive explanation to that could be a subset-minimal subset $\KB'$ of $\KB$, i.e.\ a minimal set of rules and facts, that together with $i$ \emph{suffices} to infer $o$: $\KB' \cup \{ i \} \vdash o$. 
Optionally, the explanation could contain a trace of the derivation, e.g.\ which rule fired when and which variable substitutions took place where. 
\end{example}

In MR, it is intuitive to think of attributive explanations as parts of the knowledge (represented via a knowledge base or a model) about the problem that are sufficient and/or necessary for \AI\ to yield $o$ given $i$. 
Normally, if applicable, such explanations would be required to be minimal (for some form of minimality, e.g.\ subset inclusion as in Example~\ref{ex:attributive}), consistent (for some notion of consistency or non-contradiction) and informative (or non-trivial) -- see Section~\ref{subsec:Inference} and e.g.~\cite{Falappa:Kern-Isberner:Simari:2002} for formalization in logical terms. 
Likewise, if applicable, attributive explanations may provide a trace of the attribution/association, e.g.\ inference rules, proof or some sort of argument for $o$ given $i$. 

In the case of ML, immediate examples of attributive explanations are feature attribution methods for explaining ML classifiers, such as the well-known LIME~\cite{Ribeiro:Singh:Guestrin:2016} and SHAP~\cite{Lundberg:Lee:2017}. 
There, an attributive explanation can be some minimal part $\hat{i}$ of instance $i$ that suffices for $f$ to yield $o$ (irrespective of the rest of $i$). 
Such explanations are largely underpinned by correlation between inputs $i$ and outputs $o$. 
In general, however, attributive explanations need not be only correlation-based. 
In particular, in MR they are often logical, causal, etc. 
For instance in \nameref{subsec:Planning}, an attributive explanation could exhibit (the relevant part of) the model of the problem or the state the model is in, given $i$, together with relational or causal links to $o$. 
We will see many examples of logic-based attributive explanations in Section~\ref{sec:Branches}. 
Even MR-based attributive explanations for ML models can be logical: see Section~\ref{subsubsec:implicants} and also~\cite{Ignatiev:2020-IJCAI} for examples of attributive explanations for ML classifications formalized in logical terms.

Attributive explanations are easy to design and are therefore commonplace~\cite[p.1483]{Gonul:Onkal:Lawrence:2006}. 
They are prevalent in all MR branches we consider in this report, perhaps most prominently in \nameref{subsec:Inference} (Section \ref{subsec:Inference}). 
That they are prominent among logical inference-based approaches to explainability is not surprising because, as we have seen, explainability started with early AI systems which were largely based on various classical and non-classical logics. 
That attributive explanations are prevalent in all MR approaches to explainability is not surprising either, because attribution can be seen to be necessary for building the more advanced contrastive and actionable explanations~\cite{Mohseni:Zarei:Ragan:2020}.

\subsubsection{Contrastive Explanations}
\label{subsubsec:contrastive}

\emph{Contrastive explanations} (see e.g.~\cite{Johnson:Johnson:1993,Sormo:Cassens:Aamodt:2005,Miller:2019,Sokol:Flach:2018,Krarup.et.al:2019,Mohseni:Zarei:Ragan:2020} for conceptual overviews) pertain to the notions such as criticism, contrast, counterfactual and dialogue. 
At a high level, they aim to answer the following types of questions:
\begin{enumerate}[resume*=questions]
    \item Given a representation of the problem and its solution or an answer to a query, why is something else not a solution or an answer? 
    \label{why}
    \item Given different information or query, e.g.\ a different decision or observation, would the solution/answer change too, and how?
    \label{what if}
\end{enumerate}

Questions of types~\ref{why} and~\ref{what if} pertain to reasons against the current outcome and in favor of alternative outcomes as well as to alternative inputs and parameters (see e.g.~\cite{Neches:Swartout:Moore:1985,Lacave:Diez:2002,Lim:Dey:2010,Bussone.et.al:2015,Fox:Long:Magazzeni:2017,Mohseni:Zarei:Ragan:2020}).
Such questions solicit characterization of ``the reasons for differences between a model prediction and the user’s expected outcome."\cite{Mohseni:Zarei:Ragan:2020}
Answers to~\ref{why} and~\ref{what if} types of questions thus ``inform users why an alternative output value was not produced given the current input values" and ``could provide users with enough information to achieve the alternative output value"\cite[p.~167]{Bussone.et.al:2015}. 

With our notation, contrastive explanations answer the following questions: 
\begin{quote}
    What are the details and reasons of system \AI\ yielding outcome $o$ rather than different outcome $o'$, given information $i$? \\
    And supposing that the given information is not $i$ (but some different $i'$), would $o'$ be the outcome?
\end{quote}
Contrastive explanations thus address potential criticisms of the yielded outcome $o$ given information $i$ by dealing with contrastive outcomes $o'$ and information $i'$. 

\begin{example}
\label{ex:contrastive}
Consider the planning problem mentioned in Example~\ref{ex:5G}: 
given a network reconfiguration specification, a planner's task is to produce a plan (i.e.\ a sequence of actions) of how to optimally relocate network resources. 
Let us assume that the planner operates a model $\M$ of the problem and presented with description $i$ of the initial state (i.e.\ current network configuration) and the goal state (desired configuration) yields a plan $o = \pi$ as to how to reroute the network slice. 
One may ask the question as to why a particular action $a$, say of rerouting through a specific data centre, is in the plan: $a \in \pi$ (in contrast to a plan that would not involve $a$ -- a so-called \emph{foil}~\cite{Miller:2019,Sreedharan:Srivastava:Kambhampati:2018-IJCAI}). 
A contrastive explanation to that could amount to exhibiting an exemplary plan $\pi'$ with $a \not\in \pi$, if possible, and showing that $\pi$ is better than $\pi'$ with respect to some desirable metric. 
Such a restriction $a \not\in \pi$, or more generally some plan property $P$ such as rerouting some parts before others, can also be considered as a modification of information $i$, whence the question asks how $\pi$ would change if the property had to be achieved. 
A contrastive explanation to that could amount to exhibiting other plan properties that would (not) be achieved in the contrastive case. 
Alternatively, one could ask why a contrasting plan $\pi'$ (a foil) to begin with is not a solution, to which an explanation could amount to exhibiting properties or goals that are achieved by $\pi$ but would not be achieved by $\pi'$.
\end{example}

On the one hand, contrastive explanations can work via counterfactuals~\cite{Ginsberg:1986,Wachter:Mittelstadt:Russell:2018,Byrne:2019}. 
Taken plainly, a “counterfactual is a statement such as, ‘if $p$, then $q$,’ where the premise $p$ is either known or expected to be false.”\cite[p.~35]{Ginsberg:1986} 
In contrast to attributive explanations and counterexamples, “counterfactuals continue functioning in an end-to-end integrated approach”\cite[p.~850]{Wachter:Mittelstadt:Russell:2018}, indicating consequences of changing the given information. 
So counterfactual contrastive explanations are about making or imagining different choices and analyzing what could happen or could have happened. 
For instance in \nameref{subsec:Planning}, a given difficult goal can be reduced to a sub-goal by considering counterfactual “if only thus-and-so were true, I would be able to solve the original problem” as a contrastive explanation, which then entails “arranging for thus-and-so to be true”\cite[p.~36]{Ginsberg:1986}, if possible. 
Often, counterfactual contrastive explanations address changes with respect to more desirable outcomes than the one yielded by the AI system~\cite{Ginsberg:1986,Wachter:Mittelstadt:Russell:2018,Byrne:2019}. 

In our terms, where $i$ is information at hand, a counterfactual invites one to consider what happens if different (and thus currently false) information $i'$ were at hand, speculating that it may lead to a different, perhaps more desirable, outcome $o'$. 
Taken together with the system \AI\ yielding outcome $o$ given information $i$, this expresses that `if $i'$ rather than $i$ was given, then $o'$ rather than $o$ would be the outcome'. 
For concreteness, we could write this in terms of inferences from a knowledge base: a contrastive explanation as a counterfactual can for instance be a \emph{modification} $i'$ of $i$ such that $\KB \cup \{ i' \} \models o'$. 
In addition, an attributive explanation could be incorporated by, for instance, exhibiting some minimal $\KB' \subseteq \KB$ such that $\KB' \cup \{ i' \} \models o'$, so that the counterfactual explanation indicates some minimal modification of the given information which together with some background knowledge suffices to yield a more desirable outcome. 
It may also be reasonable to strive for some minimal modification $i'$ that is in some sense most similar to $i$. 
Similarly, in ML terms, if at least part $\hat{i}$ of instance $i$ needs to be changed for $f$ to yield $o'$ instead of $o$, then a contrastive explanation as a counterfactual can be some minimal modification $i'$ of $i$ (if it exists) that necessitates $f$ to yield $o'$. 
Note though that it may be that no such modification is achievable or desirable, perhaps due to restrictions placed by the underlying application. 

On the other hand, contrastive explanations can be seen as extending and complementing attributive explanations, in that while attributive explanations provide reasons why \AI\ yields $o$ given $i$, contrastive explanations can provide counterexamples attributing to (parts of) \AI\ and $i$ why $o' \neq o$ is not possible. 
In forms of ML, a contrastive explanation as a counterexample can be some minimal part $\hat{i}$ of instance $i$ that prevents $f$ from yielding $o$ (irrespective of the rest of $i$ without $\hat{i}$) -- see ~\cite{Ignatiev.et.al:2019-NeurIPS} for examples of such contrastive explanations as counterexamples for ML classification, formalized in logical terms. 
In forms of MR, specifically in \nameref{subsec:LP} (Section~\ref{subsec:LP}) and \nameref{subsec:Argumentation} (Section~\ref{subsec:Argumentation}), in addition to attributive explanations exhibiting parts of the knowledge---a logic program or an argument graph---that make a literal true/an argument accepted, 
contrastive explanations exhibit parts of the knowledge that are conflicting and why the conflict resolution cannot result into making the literal false/argument rejected. 
Such contrastive explanations often take form in graph-like representations of pros and cons of reasoning outcomes. 
These amount to defining a formal structure, which can usually be represented as a graph, that consists of information most relevant for yielding the outcome $o$ given $i$, together with relationships revealing information dependencies and/or which information is in favor or against $o$. 
Particularly in \nameref{subsubsec:ASP} (Section~\ref{subsubsec:ASP}), graph-like contrastive explanations can be comprised of the (positive and negative) literals and rules considered in deriving the answer $o$ to the query $i$ from a logic program. 
In \nameref{subsec:Argumentation}, graphs capture the relevant supporting and conflicting information that allows to contrast the justifications of, and counterexamples to, the yielded outcome. 
Such graph-based explanations provide basis for dialogical explanations that formalize criticism and defense of the yielded outcome. 
Just as counterfactual explanations, dialogical ones allow the explainee to be engaged, rather than simply presented, with explanations. 
 
Contrastive explanations of these and similar various forms appear in \nameref{subsec:CP}, \nameref{subsubsec:ASP}, \nameref{subsec:Argumentation}, \nameref{subsec:Planning}, among others. 
They are non-trivial to define and design, given the usually numerous contrastive situations, alternative answers and solutions. 
Overall, contrastive explanations, especially counterfactual and dialogical ones, are strongly related to actionable explanations, in that contrastive explanations can support provision of actions and guidelines following which the AI system will yield a desired outcome.

\subsubsection{Actionable Explanations}
\label{subsubsec:actionable}

We maintain that \emph{actionable explanations} (see e.g.~\cite{Kulesza.et.al:2015,Bansal.et.al:2018,Cyras.et.al:2020-AAMAS,Sokol:Flach:2020,Poyiadzi.et.al:2020} for conceptual suggestions/expositions of actionability) should be interventional, interactive, collaborative, pedagogic. 
At a high level, they aim to answer the following type of question:
\begin{enumerate}[resume*=questions]
    \item Given a representation of the problem and its current solution, and given a representation of a desired outcome, what decisions or actions can be taken to improve the current solution and to achieve an outcome as close as possible to the desired one?
    \label{how}
\end{enumerate}

Questions of type \ref{how} pertain to changes to the system, modelling of or the problem itself that would lead to user-desired outputs~\cite{Lim:Dey:2010,Mohseni:Zarei:Ragan:2020}.
Such questions aim at soliciting ``hypothetical adjustments to the input or model that would result in a different output"\cite{Mohseni:Zarei:Ragan:2020}. 
Answers to~\ref{how} type of questions thus provide guidelines that help the user to achieve a (more) desired outcome. 

With our notation, actionable explanations answer the following questions: 
\begin{quote}
    What can be done in order for system \AI\ to yield outcome $o$, given information $i$? 
\end{quote}
Actionable explanations address potential interventions that may yield a desired outcome $o$, given information $i$. 
They entail both interaction and collaboration between the system and its user in that actionable explanations guide or teach the user on what actions/changes can be taken/made and the user may choose to follow them or not to. 
Importantly, actionable explanations allow to take actions or make changes that alter the system and possibly the problem themselves.

\begin{example}
\label{ex:actionable}
Consider the network reconfiguration problem mentioned in Example~\ref{ex:5G}: 
given that a congested router port is a root cause of the predicted latency violation, 
a constraint solver is tasked with finding a new routing path that avoids the congested port and best satisfies all the network slice and other operator constraints. 
Let us assume that the solver operates a collection $B \cup H$ of constraints, with $B$ the slice constraints that must be satisfied, such as the latency requirement, and $H$ the operator constraints, such as a desired cap on the number of network hops, that can be relaxed if needed. 
Suppose the solver initially finds the constraints mutually unsatisfiable. 
This can be accompanied by an attributive explanation to begin with, such as a susbset-minimal $U \subseteq H$ such that $B \cup U$ is inconsistent. 
The question then is what can be done to satisfy the constraints as best as possible while ensuring that all the slice constraints in $B$ are satisfied. 
An actionable explanation to that could be a relaxation of the operator constraints, e.g.\ allowing more network hops, that would restore satisfiability. 
For instance, a subset-minimal set $C \subseteq H$ of constraints relaxing which leads to a solution (e.g.\ $B \cup (H \setminus C)$ is consistent). 
\end{example}

So we maintain that an explanation is actionable if it suggests realizable changes to the AI system \AI\ which allow to achieve a desired outcome $o$ given information $i$. 
In MR terms for instance, an actionable explanation can be a modification $\KB'$ of constraints $\KB$ such that $\KB' \cup \{ i \} \models o$, or a modification of an argument graph $\graph$ that makes an argument of interest accepted, provided that the modification concerns the parts agreed in advance to be reasonable to modify, such as so-called \emph{soft} (or user) constraints, or the arguments that can be added/removed. 
As with contrastive explanations, it would be normal to require some form of minimality of the modification as well as some form of its similarity to the initial knowledge.\footnote{Note though that minimality may be a complicated aspect especially with respect to non-monotonic entailment in non-classical logics.}
In forms of ML, an actionable explanation can be some designation of the model’s $f$ parameter changes that result in a modified model $f'$ such that $f'(i) = o$. 
(This can be achieved e.g.\ by directly modifying the weights, or the classification thresholds or even the training set, see e.g.~\cite{Kulesza.et.al:2015} for examples of such actionable explanations for Naive Bayes Classifiers and~\cite{Lertvittayakumjorn:Specia:Toni:2020} for layer-wise relevance propagation). 
In planning or scheduling, an actionable explanation can be some minimal change of goals or resources---i.e.\ modification of the problem and hence the solver’s model---that are needed to attain a (solvable problem and its) solution satisfying as much as possible the initial goals and constraints. 

Ideally, actionable explanations would enable a meaningful interaction between an AI system and its user (human or machine, indifferently) leading to a fruitful collaboration. 
Such an interaction could be for instance conversational, formalized as dialogue between the user and the system, see e.g.~\cite{Moore:Swartout:1990,Johnson:Johnson:1993,Lacave:Diez:2002,Moulin.et.al:2002,Walton:2004,Bex:Walton:2016,Hecham:2017,Sklar:Azhar:2018,Sokol:Flach:2018,Miller:2019,Cyras.et.al:2019-ESWA,Mittelstadt:Russell:Wachter:2019,Sokol:Flach:2020-KI,Cocarascu.et.al:2020,Kraus.et.al:2020}. 
Perhaps due to their inherent complexity of dealing with arguably more consequential changes as well as attributions and contrasts, actionable explanations are not yet very prominent in MR.

\section{Explanations in MR}
\label{sec:Branches}

We here overview several classes of MR approaches where explainability is studied.

\subsection{Inference-based Explanations}
\label{subsec:Inference}

Arguably the most well established and far-back dating MR explainability techniques rest on logic-based inference (or derivation, deduction, entailment). 
\citeauthor{Falappa:Kern-Isberner:Simari:2002} outline well in~\cite{Falappa:Kern-Isberner:Simari:2002} the notion of explanation used in logic-based reasoning formalisms as follows. 
Assuming an underlying logic and an appropriate formalization of derivation ($\vdash$), (in)consistency ($\bot$) and logical consequence ($Cn$), an explanation for a sentence $a$ is a minimal, consistent and informative set $A$ of sentences deriving $a$:
\begin{align}
    A \vdash a, ~ A \nvdash \bot, ~ B \nvdash a \text{ for } B \subsetneq A, ~ Cn(A) \nsubseteq Cn(a).
\label{eqn:inference}
\end{align}
Such and similar \inference\ explanations in logic-based reasoning are generally non-unique, so that preferred explanations are often selected using some specific ordering criteria~\cite{Pino-Perez:Uzcategui:2003} with various forms of minimality (e.g.\ subset inclusion, cardinality, depth of proof/inference) being a frequent choice. 
In addition, \inference\ explanations are often accompanied by a trace or a proof tree of the inference~\cite{Ferrand.et.al:2005} -- the Cyc project is a well-known example of an MR-based AI system providing \inference\ explanations with traces~\cite{Baxter.et.al:2005}. 
Overall, \inference\ explanations are \textbf{attributive} in nature and aim to answer the following kind of question:
\begin{itemize}
    \item What explains a given observation? In other words, which information logically entails a given inference?
\end{itemize}
We next briefly overview some prominent examples of \inference\ explanations used in MR today, with applications to e.g.\ ML.

\subsubsection{Axiom Pinpointing}
\label{subsubsec:pinpointing}

In description logics (which are decidable fragments of first-order logic), axiom pinpointing~\cite{Penaloza:Sertkaya:2017} is a perfect example of the well established inference-based explanation concepts still being very relevant to modern MR explainability. 
Axiom pinpointing amounts to finding axioms in a knowledge base $\KB$ that entail or prevent a given consequence/query $Q$ (e.g.\ a Boolean conjunctive), whereby minimal such sets of axioms are taken as justifications/explanations: 
roughly, given $\KB \models Q$, an explanation is a minimal $\KB' \subseteq \KB$ with $\KB' \models Q$.
(The case where $\KB \not\models Q$ is more nuanced, but an explanation roughly amounts to $\KB' \subseteq \KB$ that is inconsistent with $Q$~\cite{Bienvenu.et.al:2019}.) 
Such \textbf{attributive} explanations aim to answer the following question~\cite{Arioua:Tamani:Croitoru:2015,Bienvenu.et.al:2019}:
\begin{itemize}
    \item Given a knowledge base $\KB$ and a query $Q$, why is $Q$ (not) entailed by $\KB$?
\end{itemize}
Works of \cite{Kalyanpur.et.al:2007,Horridge:Parsia:Sattler:2009,Bienvenu.et.al:2019,Ceylan.et.al:2019,Lukasiewicz.et.al:2020} are examples of explaining knowledge base query answering where explanations are defined as (minimal) subsets entailing or contradicting a given query with respect to a (consistent or inconsistent) knowledge base.

\subsubsection{Implicants}
\label{subsubsec:implicants}

In the research strand called Model Diagnosis~\cite{Reiter:1987,Darwiche:1998}, what can now be seen as explanations used to be called \emph{diagnoses} for a faulty system (encoded as a logical model). 
These were defined via (\emph{prime}) \emph{implicants} (see e.g.~\cite[Defn.~6, p.~171]{Darwiche:1998} for a definition) -- essentially inference-based explanations as in Equation~\ref{eqn:inference}. 
Some modern MR works use prime implicants for explainability of, notably, ML models. 
There, the idea is roughly that, using a logical encoding of an ML classifier, \textbf{attributive} inference-based explanations correspond to subsets of features of instances that are sufficient to yield the predicted classes.

For example, in~\cite{Shih:Choi:Darwiche:2018,Darwiche:Hirth:2020} the authors assume a Bayesian network classifier to be encoded as a propositional logic formula $\Delta$ so that classification corresponds exactly to propositional entailment. 
In particular, an input instance $\alpha$ to the classifier as a set of features corresponds to a conjunction of literals, and the output (i.e.\ classification of $\alpha$) corresponds to entailment of $\Delta$ 
(e.g.\ $\alpha \models \Delta$ or $\alpha \models \neg\Delta$, in case of binary classifiction). 
Then, sub-formulas of instances (i.e.\ subsets of features) that are prime implicants of $\Delta$ (or $\neg\Delta$) are seen as explanations of classifications. 

In a similar line of work~\cite{Ignatiev.et.al:2019,Ignatiev.et.al:2019-NeurIPS}, the authors exploit prime implicants of either propositional or first-order logic formulas to explain predictions of ML classifiers that can be logically encoded as sets of constraints. 
In contrast to~\cite{Shih:Choi:Darwiche:2018,Darwiche:Hirth:2020}, \citeauthor{Ignatiev.et.al:2019-NeurIPS} also provide \textbf{contrastive} explanations via counterexamples to classifications~\cite{Ignatiev.et.al:2019-NeurIPS}.

\subsubsection{Abduction}
\label{subsubsec:abduction}

Inference-based explanations in early MR have been widely studied in abductive reasoning,  e.g.~\cite{Shanahan:1989,Marquis:1991,Kakas:Kowalski:Toni:1992,Eiter:Gottlob:1995,Leake:1995,Kohlas:Berzati:Haenni:2002,Torta.et.al:2014}. 
Abduction aims at explaining observations by means of inventing, i.e.\ abducing, additional knowledge that allows to infer those observations. 
The abduced knowledge, possibly together with the existing background knowledge, is broadly seen as an explanation of an observation as follows. 
In a logic-based setting, assume a background knowledge $\KB$, a (possibly hypothetical) observation $o$ and a space $H$ of hypotheses (abducibles). 
Then, a set $ab \subseteq H$ of abducibles is an explanation for $o$ just in case $\KB \cup ab \models o$. 
(Consult \cite{Eiter:Gottlob:1995} for an excellent overview of variants of abduction, including criteria of consistency, minimality and prioritisation.) 

In modern day AI, \textbf{attributive} abductive explanations can be used in different settings. 
In MR, for instance, they can help to explain abductive reasoning itself, e.g.\ via causal knowledge graphs as in~\cite{Torta.et.al:2014}. 
In ML, they can (via prime implicants) be used for explaining classifications of logically encoded ML models by abducing minimal assignments to variables (which represent features) that guarantee classifications, as in~\cite{Ignatiev.et.al:2019-NeurIPS,Bertossi:2020}.

\medskip

Abduction in general is a form of non-monotonic reasoning~\cite{Eiter:Gottlob:1995}. 
That is, expanding the background knowledge may shrink the availability of explanations: for $\KB' \supsetneq \KB$, the previous explanation $ab$ for $o$ with respect to $\KB$ may no longer be an explanation with respect to $\KB'$ (because e.g.\ $\KB' \cup ab \models \bot$). 
It is thus not surprising that inference-based explanations appeared early in default reasoning~\cite{Poole:1985} and other non-classical logic-based approaches, notably Logic Programming. 
We overview explainability in the latter next.

\subsection{Logic Programming (LP)}
\label{subsec:LP}

Logic Programming (LP, see e.g.~\cite{Apt:Bol:1994,Kowalski:2014}) is both a knowledge representation formalism and computational mechanism for non-classical, particularly non-monotonic, reasoning. Often, explainability in LP is enabled by the declarative reading of rules in logic programs, which allows for \textbf{attributive} explanations in terms of knowledge and rules that yield a specific inference. 
Further, if the knowledge and rules carry immediate semantic meaning, then deductive inference paths or proof trees are readily interpretable and can be translated into natural language, as e.g.\ in Rulelog~\cite{Rulelog:2017,Rulelog:2018}. 
Still further, \textbf{contrastive} explanations can be obtained by inspecting conflict resolution strategies. 

In general, LP takes various forms, most notably abductive, inductive and answer set programming, each with different computational procedures and different approaches to explainability. 
We discuss some of these below.

\subsubsection{Abductive Logic Programming (ALP)}
\label{subsubsec:ALP}

Abductive Logic Programming (ALP)~\cite{Eshghi:Kowalski:1989,Kakas:Kowalski:Toni:1992} is a form of abductive reasoning expressed using LP vocabulary. Abductive logic programs are used for knowledge representation and abductive proof procedures for automated reasoning. Abductive proof procedures interleave backward and forward reasoning and can be used for checking and enforcing properties of knowledge representation (via queries or inputs to the program) as well as for agents to abduce and/or explain actions required to check/enforce such properties. 
In a nutshell, given an abductive logic program $P$, an abductive explanation for an observation $o$ (conjunctive formula) is a pair $(ab, \theta)$ consisting of a (possibly empty) set $ab \subseteq H$ of abducibles from a set $H$ of variable-free atoms (i.e.\ hypotheses) and a (possibly empty) variable substitution $\theta$ for the variables in $o$ such that $P$ together with abducibles $ab$ entail $o\theta$ and satisfy integrity constraints. 
Different notions of entailment and satisfaction can be adopted, for example classical first-order entailment $P \cup ab \models o\theta$ and consistency $P \cup ab \nvdash \bot$. 

Intuitively, abductive explanations answer the following questions~\cite{Kakas:Kowalski:Toni:1992,Toni:2001}:
\begin{itemize}
    \item Why did this observation occur? 
    \item What explains this observation? 
    \item How to reach this goal/query? 
\end{itemize}
Accordingly, as in Section~\ref{subsubsec:abduction} \nameref{subsubsec:abduction}, abductive explanations in ALP are \textbf{attributive}. 
However, they can be seen to have a flavour of actionability:
“explanations can be thought of as data to be actually added to the beliefs of the agents and actions that, if successfully performed by the agents, would allow for the agents to achieve the given objectives while taking into account the agents' beliefs, prohibitions, obligations, rules of behavior and so on, in the circumstances given by the current inputs.”\cite[p.~100]{Toni:2001} 
In other words, abductive explanations suggest how to modify the knowledge (represented e.g.\ by a logic program) so that the objective (e.g.\ observation, query) obtains. 
In this flavour, ALP explanations have been related~\cite{Wakaki:Nitta:Sawamura:2009,Booth.et.al:2014,Sakama:2018} to other forms of explanations in MR, particularly in Argumentation (see Section~\ref{subsec:Argumentation}).

\subsubsection{Inductive Logic Programming (ILP)}
\label{subsubsec:ILP}

A complementary LP technique to abduction is that of induction~\cite{Muggleton:Raedt:1994}: “abduction is the process of explanation -- reasoning from effects to possible causes, whereas induction is the process of generalization -- reasoning from specific cases to general hypothesis.”\cite[p.~205]{Law:Russo:Broda:2019}
Inductive Logic Programming (ILP)~\cite{Muggleton:Raedt:1994} studies the inductive construction of logic programs from examples and background knowledge. 
Briefly, “given a set of positive examples, and a set of negative examples, an ILP system constructs a logic program that entails all the positive examples but does not entail any of the negative examples.”\cite[p.~1]{Evans:Grefenstette:2018} 
The set of induced rules, called hypotheses, possibly together with a proof trace, is viewed as an explanation of the examples in the context of the background knowledge:
given background knowledge $\KB$ and sets $P$ and $N$ of positive and negative examples, respectively, a set $R$ of clauses (i.e.\ hypotheses) is an explanation for given examples just in case $\KB \cup R \vdash p$ for all $p \in P$ and $\KB \cup R \not\vdash n$ for all $n \in N$. 
Such explanations are \textbf{attributive} in nature and aim to answer the following question:
\begin{itemize}
    \item What general hypothesis best explains the given specific examples/observations?
\end{itemize}
Recent applications of ILP for explainability include integrating ILP and ML techniques to learn explanatory logic programs from non-symbolic data~\cite{Evans:Grefenstette:2018,Minervini.et.al:2020}.

\subsubsection{Answer Set Programming (ASP)}
\label{subsubsec:ASP}

Answer Set Programming (ASP)~\cite{Brewka:Either:Truszczynski:2011,Lifschitz:2019} is a formalism for default reasoning with knowledge represented in LP vocabulary. 
Explanations in ASP have been studied for some 25 years and \citeauthor{Fandinno:Schulz:2019} provide an excellent overview in~\cite{Fandinno:Schulz:2019}. 
There are two main families of approaches, namely justification and debugging. 
The respectively aim at answering the following questions:
\begin{itemize}
    \item Why is a literal (not) contained in an answer set?
    \item Why is an unexpected or no answer set computed?
\end{itemize} 
Both families first-and-foremost provide \textbf{attributive} explanations, albeit with different flavours: 
justifications can be inspired by, for instance, causal or argumentative reasoning; 
debugging can be based, for instance, on reporting unsatisfied rules or unsatisfiable cores. 
Both justification and debugging approaches may also supply \textbf{contrastive} explanations by including conflicting information and revealing conflict resolution that takes place in reasoning. 
We invite the reader to consult~\cite{Fandinno:Schulz:2019} for a detailed exposition, but briefly summarize the main ideas below, and afterwards discuss some most recent works.

\paragraph{Justification approaches}

Justification approaches by and large concern consistent logic programs and provide “somewhat minimal explanation as to why a literal in question belongs to an answer set.”\cite[p.~119]{Fandinno:Schulz:2019} 
\emph{Off-line justifications}~\cite{Pontelli.et.al:2009} are graph structures describing the derivations of atoms’ truth values via program rules and can be seen to provide traces of dependencies. 
\emph{Labelled Assumption-Based Argumentation (ABA)-based answer set justifications} (LABAS)~\cite{Schulz:Toni:2013,Schulz:Toni:2016} abstract away from intermediate rule applications and focus on the literals occurring in rules used in the derivation and can be seen to provide traces via (supporting and attacking) arguments (see Section~\ref{subsec:Argumentation}~\nameref{subsec:Argumentation}), thus exhibiting reasons pro and con the inference in question. 

In a similar vain, \emph{causal graph justifications}~\cite{Cabalar:Fandinno:Fink:2014} associate with each literal a set of causal justifications that can be graphically depicted and can be seen as causal chain traces. 
Causal graph justifications are inspired by
\emph{why-not provenance}~\cite{Damasio:Analyti:Antoniou:2013}, which itself provides non-graphical justifications expressing modifications to the program that can change the truth value of the atom in question. 
(By extension, justifications for the actual truth values do not imply any modifications.) 
Why-not provenance justifications can be seen to approach the realm of actionable explanations, though unlike causal graph justifications, why-not provenance does not discriminate between productive causes and other counterfactual dependencies (see e.g.~\cite{Hall:2004,Hall:2007}). 

\paragraph{Debugging approaches}

Unlike justification approaches, debugging approaches by and large concern inconsistent logic programs and provide explanations as to why a set of literals is not an answer set. 
The \emph{spock system}~\cite{Brain:Vos:2008,Gebser.et.al:2008} transforms a logic program into a meta-(logic)program, expressing conditions for e.g.\ rule applicability, in such a way that the answer sets of the meta-program capture violations, in terms of rule satisfaction, unsupported atoms or unfounded loops, of the candidate answer set of the original program. 
The \emph{Ouroboros} system~\cite{Oetsch.et.al:2010} extends these ideas to logic programs possibly with variables. 
The authors there also tackle the issue of multiple explanations by requiring the user to specify an intended answer set. 
To avoid the need to specify a whole answer set to begin with, the authors of~\cite{Shchekotykhin:2015,Dodaro.et.al:2019} instead propose \emph{interactive debugging} whereby the user is queried about specific atoms in order to produce the relevant explanations of inconsistencies. 

\paragraph{Recent applications}

The above summary shows the wealth of explainability techniques in ASP, but there are also works that use ASP for explanations in other areas. 
(These more recent works are not overviewed in~\cite{Fandinno:Schulz:2019}.) 
For instance, abstraction in ASP~\cite{Saribatur:2020} can be used for explainability in a number of ASP-supported problem solving tasks, including problems in scheduling and planning~\cite{Saribatur:2020-XLoKR}. 
ASP can also complement ML techniques in classification and explanation. 
For instance, 
in~\cite{Calegari.et.al:2019-WOA}~\citeauthor{Calegari.et.al:2019-WOA} map decision trees (DTs) into logic programs to explain DT predictions via natural language explanations generated from LP rules. 
In~\cite{Riley:Sridharan:2019}~\citeauthor{Riley:Sridharan:2019} integrate ASP-based reasoning and learning with DTs and deep learning for visual question answering. 

In another recent research direction~\cite{Bertossi:2020}, \citeauthor{Bertossi:2020} uses ASP to generate \textbf{contrastive} explanations for discrete, structured data classification. 
Briefly, \citeauthor{Bertossi:2020} proposes the following explanations: 
\begin{enumerate*}[a)]
    \item causal explanations are sets of feature-value pairs such that changing at least one value changes the classification label; 
    \item counterfactual (value-)explanations are individual feature-value pairs such that changing the value changes the classification; 
    \item actual (value-)explanations are individual feature-value pairs together with a set of feature-value pairs such that changing the value of the former does not suffice to change the classification, but changing the values of both the former and the latter does. 
\end{enumerate*}
Actual and counterfactual explanations can be assigned explanatory responsibility measure, which amounts to the inverted size of the smallest set accompanying an actual explanation, with the explanatory responsibility of a counterfactual explanation always being 1 (because it suffices to flip only that feature-value to change the classification). 
Overall, \citeauthor{Bertossi:2020} proposes to encode the inputs-outputs of an ML classifier into an ASP program, together with predicates and constraints capturing interventions (i.e.\ feature-value flipping) to extract the various explanations defined by computing answer sets.

\subsection{Constraint Programming (CP)}
\label{subsec:CP}

Constraint Programming (CP)~\cite{CP:2006} is a paradigm for solving combinatorial search problems, often called Constraint Satisfaction Problems (CSPs). 
CSPs are represented in terms of decision variables and constraints, usually in classical logic vocabulary, and solving them amounts to finding value assignments to variables that satisfy all the constraints.\footnote{We note that it is also very natural to use LP (see Section~\ref{subsec:LP}), particularly ASP (Section~\ref{subsubsec:ASP}), in solving CSPs.}

As with the MR techniques described in Sections~\ref{subsec:Inference} and~\ref{subsec:LP}, 
inference is a critical part of constraint solving and was used early for explainability. 
For instance, in~\cite{Sqalli:Freuder:1996}, a “trace of the inference, with some rudimentary natural language processing, provided explanations for puzzles taken from newsstand puzzle booklets that were reasonably similar to the answer explanations provided in the back of the booklets.”\cite[p.~4860]{Freuder:2017}  
However, already in mid-90s it was understood that explanations in CP cannot simply amount to tracing of solutions but need to be more sophisticated. 
To quote from~\cite[p.~4860]{Freuder:2017}:
\begin{quote}
  A natural approach to providing a richer explanation of a solution would be to ‘trace’ the program’s solution process. However, constraint solvers generally employ search, and tracing search tends not to provide a very satisfying explanation. For backtrack search: “I tried this and then that and hit a dead end, so I tried the other instead”. Even worse, for local search: “I kept getting better, but then I tried some other random thing”. 
\end{quote}
Hitting a dead end, or a failure, usually amounts to finding the constraints and variable assignments considered so far unsatisfiable. 
Identification of mutually unsatisfiable constraints is indeed a key aspect of both solving and explaining CSPs. 
For concreteness, we next discuss how unsatisfiability relates to explainability in SAT (Boolean satisfiability) -- a sub-branch of CP.

\subsubsection{SAT and Beyond}
\label{subsubsec:SAT}

A SAT problem~\cite{Davis:Putnam:1960} is defined as follows. 
Given a Boolean formula $f(x_{0},\ldots,x_{n})$, check if $f$ is satisfiable, i.e.\ if there is an assignment to variables $x_{0}$ through $x_{n}$ which renders $f$ true. 
SAT problems form a special class of CSPs. 
There can be many variants of SAT, obtained by placing various constraints, either on the structure of the formula, or on the class of allowable Boolean operators. 
Generalizing SAT by replacing sets of variables with predicates from various first-order logic theories yields the class of Satisfiability Modulo Theories (SMT) problems. 
These are highly expressive and, along SAT, widely used in applications. 

\paragraph{Minimal Unsatisfiable Sets (MUSes)}
\label{par:MUS}

One main challenge in SAT, SMT and CP overall is that of explaining failure (also called infeasibility, unsolvability, unsatisfiability, inconsistency). 
In SAT, an explanation of unsatisfiability of formula $f$ is commonly defined as a minimal unsatisfiable sub-formula (words `core' and `set' can be used interchangeably with sub-formula).
Concretely, following~\cite{Previti:Marques-Silva:2013}, assume $f$ to be in conjunctive form, i.e.\ a conjunction of clauses (i.e.\ disjunctions of literals). 
Note that $f$ can also be seen as a set of clauses and each clause as a set of literals. 
Then, an explanation for why $f$ is unsatisfiable is a \emph{minimal unsatisfiable set} (MUS) of $f$, 
i.e.\ $U \subseteq f$ such that $U \models \bot$ and for any $U' \subsetneq U$ we have $U' \not\models \bot$. 

\begin{example}
\label{ex:MUS}
Let 
$f(a, b, c) = \stackrel{C_1}{(a)} \land \stackrel{C_2}{(\lnot a \lor b)} \land \stackrel{C_3}{( \lnot b)} \land \stackrel{C_4}{(c)}$
consist of four clauses $C_1, \ldots, C_4$ as indicated. 
$f$ is unsatisfiable and this is explained by the MUS $U = \{ C_1, C_2, C_3 \}$. 
Indeed, one can remove $C_4$ from $f$ and the resulting set $U$ would still be unsatisfiable; 
yet every pair of clauses $C_1, C_2, C_3$ is satisfiable. 
\end{example}

Intuitively, MUSes identify the culprits that generate inconsistency, regardless of the rest. 
MUSes are \textbf{attributive} explanations and answer the following question:
\begin{itemize}
    \item Which constraints are sufficient to lead to inconsistency?
\end{itemize}


There is a clear connection between MUSes and \inference\ explanations, particularly prime implicants (see Section~\ref{subsubsec:implicants}). 
Indeed, note that for a formula $g$ and literal $l$ it holds that $g \models l$ iff $g \land \lnot l \models \bot$. 
Thus, prime implicants of $l$ (from $g$) can be found by computing MUSes of $g \land \lnot l$ and removing $\lnot l$ thereof. 

\begin{example}
\label{ex:prime implicant}
Similarly to Example~\ref{ex:MUS}, consider 
$g(a, b, c) = \stackrel{C_2}{(\lnot a \lor b)} \land \stackrel{C_3}{( \lnot b)} \land \stackrel{C_4}{(c)}$ so that $g \models \lnot a$. 
Then $g \land a = f$ admits an MUS $\{ C_1, C_2, C_3 \}$, as in Example~\ref{ex:MUS}, and removing $C_1 = a$ yields the prime implicant $g' = \stackrel{C_2}{(\lnot a \lor b)} \land \stackrel{C_3}{( \lnot b)}$ of $\lnot a$.
\end{example}

Computation of prime implicants using MUSes has been exploited to e.g.\ explain classifications of ML models logically encoded using CP in~\cite{Ignatiev.et.al:2019}. 
In addition, MUSes are widely used as building blocks of explanations in other MR approaches, for instance \nameref{subsec:Planning}~ \cite{Eifler.et.al:2020-AAAI,Eifler.et.al:2020-IJCAI}, \nameref{subsec:Argumentation}~\cite{Niskanen:Jarvisalo:2020} and Decision Theory~\cite{Boixel:Endriss:2020}.

\paragraph{Minimal Correction Sets (MCSes)}
\label{par:MCS}

Once a failure obtains and is explained by potentially multiple MUSes while solving a SAT (or an SMT) problem, the main challenge is to address it by making the problem solvable (i.e.\ making the constraints feasible/satisfiable/consistent). 
This can be achieved by identifying the modifications (or relaxations) of constraints that allow to restore consistency. 
Concretely, using SAT terms as in Section~\ref{par:MUS}, an explanation as to how an unsatisfiable formula $f$ can be made satisfiable is a \emph{minimal correction set} (MCS) of $f$, 
i.e.\ $C \subseteq f$ such that $f \setminus C \not\models \bot$ and for any $C' \subsetneq C$ we have $f \setminus C' \models \bot$. 
In other words, an MCS is a minimal set of clauses that if removed from $f$ would make the resulting formula satisfiable. 

\begin{example}
\label{ex:MCS}
The explanations as to how the unsatisfiable $f$ from Example~\ref{ex:MUS} can be made satisfiable are the clauses $C_1, C_2, C_3$: 
if any of them were removed from $f$, the resulting formula would be satisfiable. 
\end{example}

MCSes can be seen as \textbf{contrastive} explanations in that they show how different collections of constraints (expressed via SAT or SMT formulas) are satisfiable in contrast to the given collection. 
MCSes thus answer the following question:
\begin{itemize}
    \item Which constraints should be relaxed in order to recover consistency?
\end{itemize}
However, note that MCSes in general do not discriminate between the constraints (clauses) that are more or less modifiable. 
This is somewhat in contrast to explanations in CP more generally, as discussed next.

\subsubsection{General CP}
\label{subsubsec:CP}

In CP, especially concerning explainability, it is common to distinguish between two types of constraints, namely
background and user (also called hard and soft, static and dynamic, etc.) constraints: 
the former, denoted $B$, represent constraints that need to be satisfied and cannot be changed; 
the latter, denoted $H$, represent constraints that are desirable to satisfy but can be modified if needed. 
Then, explanations in CP often amount to exhibiting sets of constraints that are mutually inconsistent and sets of constraints that need to be relaxed (i.e.\ dropped or changed) to restore consistency, e.g.~\cite{Amilhastre.et.al:2002,OSullivan.et.al:2007}: 
$U \subseteq H$ such that $B \cup U$ is inconsistent, and $C \subseteq H$ such that $B \cup (H \setminus C)$ is consistent. 
As with \inference\ explanations, forms of minimality or other preferences can be used to select the desirable explanations~\cite{Freuder.et.al:2002,Junker:2004}. 
Thus, the notions of MUS and MCS can obviously be lifted from SAT in particular to CP in general. 
For instance, an MUS can refer to the so-called \emph{nogood} -- a minimal inconsistent (or conflicting) set of constraints~\cite{Amilhastre.et.al:2002}; 
an MCS can refer to the so-called \emph{minimal exclusion set} -- a minimal set of constraints to remove in order to restore consistency~\cite{OSullivan.et.al:2007}. 

As with MUSes in Section~\ref{subsubsec:SAT}, the sets of constraints that are collectively unsatisfiable can be seen as \textbf{attributive} explanations. 
They answer the following questions.
\begin{itemize}
    \item Why is there a conflict between these parts of the system?~\cite{Freuder.et.al:2002}
    \item Which constraints result into failure?~\cite{Junker:2004}
    \item From which subsets of current choices did inconsistency follow?~\cite{Amilhastre.et.al:2002,Freuder:2017}
\end{itemize}
Instead, constraint relaxations can be seen as \textbf{actionable} explanations in that they show how to change the designated part (i.e.\ user constraints) of the CSP to achieve the desirable outcome of solvability. 
They answer the following questions. 
\begin{itemize}
    \item Which choices should I relax in order to recover consistency?~\cite{Amilhastre.et.al:2002,Freuder:2017}
    \item Which choices should I relax in order to render such a value available for such a variable?~\cite{Amilhastre.et.al:2002,Freuder:2017}
\end{itemize}

More generally, explanations can be defined for any inference on constraints, not only inconsistency~\cite{Freuder.et.al:2002,Downing:Feydy:Stuckey:2012}. 
Explainability then amounts to finding constraints or variable assignments\footnote{For simplicity, a variable assignment can be seen as a constraint.} that entail other constraints. 
It is thus instructive to think of CSP solving as a process where against the background constraints, user constraints are being propagated~\cite{Freuder.et.al:2002}. 
Then, at a given snapshot of the solving process, 
we can focus on a set $\KB \subseteq B \cup H$ of currently considered constraints, some constraint $i$ in consideration and the latest inference $o$ thereof. 
For example, $i$ can be the last user added constraint, a variable assignment or a background constraint of interest; 
similarly, $o$ can be a variable value restriction, e.g.\ literal $l$ set to true ($l=\top$) or integer variable $v$ forced to takes values in between 1 and 2 ($v \in [1, 2]$), or failure $\bot$. 
Explaining the inference $o$ then amounts to finding the part of currently considered constraints that suffice to entail $o$ when propagating $i$. 
Such attributive explanations aim to answer the following questions.
\begin{itemize}
    \item Why does this parameter have to have this value? \cite{Freuder.et.al:2002}
    \item What are the next propagation steps? \cite{Winter:Stuckey:Musliu:2020,Downing:Feydy:Stuckey:2012}
    \item Why is this value not available any longer for this variable? \cite{Amilhastre.et.al:2002,Freuder:2017}
\end{itemize}

Overall then, explanations in CP can be roughly described as follows. 
On the one hand, given a set $\KB$ of constraints with a constraint $i$ of interest, an \textbf{attributive} explanation for the latest inference $o$ is 
a minimal $\KB' \subseteq \KB$ such that $\KB' \cup \{ i \} \models o$ 
(assuming appropriate notions of entailment $\models$ and minimality). 
In particular, if $i \in H$ is imposed on some $\KB = B \cup H' \subseteq B \cup H \setminus \{ i \}$ and this results into inconsistency $o = \bot$, 
then a $\subseteq$-minimal $U \subseteq H'$ such that $B \cup U \cup \{ i \}$ is inconsistent is an explanation of failure.
On the other hand, given $\KB \cup \{ i \} \models \bot$,  an \textbf{actionable} explanation can be 
a modification $\KB'$ of $\KB$ such that $\KB' \cup \{ i \} \not\models \bot$. 
In particular, for $i \in H$ and $H' \subseteq H \setminus \{ i \}$, 
a relaxation $C \subseteq H'$ such that $B \cup (H' \setminus C) \cup \{ i \}$ is consistent is an explanation of how to restore consistency by relaxing constraints $C$. 

Interestingly, explanations pertaining to constraint inconsistency and relaxations are often used in CP for improving the solution strategies themselves:
“much of the work on explaining failure actually is focused on programs explaining intermediate failures to themselves in order to reach a solution more efficiently.”\cite[p.~4860]{Freuder:2017}.
We review below several approaches in CP to devising explanations conceptually similar to those delineated above. 

A notable work is that on explaining \emph{alldifferent}~\cite{Downing:Feydy:Stuckey:2012},
where explanations are produced for improving solver strategies. 
The motivation of~\citeauthor{Downing:Feydy:Stuckey:2012} is as follows~\cite[p.~116, emphasis original]{Downing:Feydy:Stuckey:2012}:
\begin{quote}
    Whenever a propagator changes a domain it must explain how the change occurred in terms of literals, that is, each literal $l$ that is made true must be explained by a clause $L \to l$ where $L$ is a (set or) conjunction of literals. When the propagator causes failure it must explain the failure as a \emph{nogood}, $L \to \bot$, with $L$ a conjunction of literals which cannot hold simultaneously.
\end{quote}
Roughly then, the propagator’s actions can be explained using cut-sets, where an explanation is effectively a logical constraint on (the values of) variables. 
Similar ideas using a lazy clause generation solver for explaining propagation via constraints from which nogoods can be computed can be applied to string edit distance constraints, e.g.\ in~\cite{Winter:Stuckey:Musliu:2020}~\citeauthor{Winter:Stuckey:Musliu:2020} use explanations that consist of literals which logically entail the truth of a Boolean variable that encodes propagation of some variable’s value. 

Apart from explainability for improving solver themselves, there are works that supply explanations for solving a CSP to the user. 
In~\cite{Amilhastre.et.al:2002}~\citeauthor{Amilhastre.et.al:2002} provide “explanations for some user’s choices and ways to restore consistency”, whereby “the user specifies her requirements by interactively giving values to variables or more generally by stating some unary constraints that restrict the possible values of the decision variables.” 
The goal is to “provide the user with explanations of the conflicts” in terms of “(minimal) inconsistent subsets of the current set of choices.” 

In his seminal paper~\cite{Junker:2004} from~\citeyear{Junker:2004},~\citeauthor{Junker:2004} proposed QuickXplain -- a general purpose technique for explainability in constraint programming. 
In a general setting of constraints, relaxations (resp.\ conflicts) are defined as sets of constraints for which (resp.\ no) solution exists. 
Conflicts thus explain solution failure, relaxations restore consistency. 
However, both are generally exponential to construct and present, whereas a user may desire explanations pertaining to the most important constraints. 
Thus, preferred relaxations and preferred conflicts are defined to be minimal with respect to lexicographic orderings (over relaxations and conflicts) defined using any total order over constraints. 
In practice, this amounts to successively adding the most preferred constraints until they fail and then removing the least preferred constraints as long as that preserves failure. 
Explainability-wise, “preferred conflicts explain why best elements cannot be added to preferred relaxations”.\cite[p.~169]{Junker:2004}\footnote{Effectively, such explanations can be seen as a form of Brewka’s preferred subtheories~\cite{Brewka:1989} in the language of constraints.} 
Further, the described “checking based methods for computing explanations work for any solver and do not require that the solver identifies its precise inferences.”\cite[p.~172]{Junker:2004} 
In a more general CP setting~\cite{OSullivan.et.al:2007} similar to~\cite{Junker:2004},~\citeauthor{OSullivan.et.al:2007} propose representative explanations (and algorithms thereof) in which “every constraint that can be satisfied is shown in a relaxation and every constraint that must be excluded is shown in an exclusion set.”\cite[p.~328]{OSullivan.et.al:2007}

In general, \citeauthor{OCallaghan.et.al:2005} argue in~\cite{OCallaghan.et.al:2005} that “desirable is an explanation that is corrective in the sense that it provides the basis for moving forward in the problem-solving process”. 
The authors there formally define a corrective explanation that intuitively is “a reassignment of a subset of the user’s unary decision constraints that enables the user to assign at least one more variable”. 
They as well provide an algorithm for computing corrective explanations of minimal length. 

Recently, there have also been works that consider a constraint solver assisting a human user in solving some logical problem. 
For instance, in~\cite{Bogaerts.et.al:2020}~\citeauthor{Bogaerts.et.al:2020} explain “the propagation of a constraint solver through a sequence of small inference steps”. 
They use minimal unsatisfiable sets of constraints for generating explanations of the solver’s individual inference steps and the explanations can overall be seen as proofs or traces of the solver’s working towards a solution. 

%
%

\subsection{Automated Theorem Proving (ATP) and Proof Assistants}
\label{subsec:ATP}

Automated Theorem Proving (ATP) is a procedure whereby a tool known as a “theorem prover” is provided a proposition, and it returns ‘true’ or ‘false’ (or runs out of time). 
Theorem provers have matured tremendously in recent years, and are now used in many settings, not only for arithmetic. In general, SAT solvers, SMT solvers, and model checkers also fall under the ambit of theorem proving. 
Proof assistants are special kinds of theorem provers (often called interactive theorem provers). These are hybrid tools that automate the more routine aspects of building proofs while depending on human guidance for more difficult aspects. 
The user can write a theory of their choice and verify whether or not a proposition holds of this theory. 
Overall, a proof assistant might return a true/false answer, run out of time, or stop at a subgoal that it does not “know” how to solve. 
In the last case, the user might need to write helper code and expand the theory.
In essence, proof assistants are interactive, that is, they often require more input from the user to solve “difficult” goals. 
Well known examples of general proof assistants are Isabelle\footnote{\href{https://isabelle.in.tum.de/}{https://isabelle.in.tum.de/}}, F*, Coq\footnote{\href{https://coq.inria.fr/}{https://coq.inria.fr/}} etc.~\cite{Bertot:Casteran:2004}. 
Specialized proof assistants exist as well, such as Tamarin for security protocol verification. 
Proof assistants are being widely used for a variety of applications, including building verified compilers~\cite{Krebber.et.al:2014} and verified implementations of processors~\cite{Pulte.et.al:2019}.

Most theorem provers provide no explanation for a proposition which is verified to be true~\cite{Fitting:1996,Sutcliffe:Suttner:2001}. 
However, some interactive theorem provers, i.e.\ proof assistants, do. 
In particular, if a proof assistant returns ‘true’, it may generate a sequence of steps that one can then use to replicate the proof by hand for better understanding. 
This can be seen as an \textbf{attributive} explanation of the prover's outcome. 
On the other hand, if a specialized proof assistant or a general theorem prover returns ‘false’, it can usualy generate a falsifying counterexample to the user as to why the proposition does not hold. 
This can be seen as a \textbf{contrastive} explanation. 
Note, however, that not all provers do that -- Coq, for instance, would just show the user some pending goals which it cannot solve (and the statements corresponding to these goals might themselves be false), or throw an error message to say that the proposition cannot be proved true by some underlying decision procedure.
In sum, explanations from theorem provers, when they are available, aim at answering the following questions.
\begin{itemize}
    \item If a proposition is true, what is a (shortest/most readable) proof?
    \item If a proposition is false, what is a counterexample?
\end{itemize}

There are obviously open problems pertaining to explainability in ATP, to name a few.
\begin{enumerate*}
    \item Can a theorem prover/proof assistant also be optimized to provide the “best” counterexample, according to some measure? 
    \item Can non-interactive theorem provers also provide explanations/proof descriptions for propositions that are verified to be true? What would these explanations look like?
    \item If a (sub)goal involves an unsolvable loop which leads to a timeout, can this be output as an explanation instead of (or in addition to) timing out?
\end{enumerate*}
Furthermore, currently the explanations generated by most theorem provers and proof assistants are not very human-friendly. Some proofs/counterexamples generated by some theorem provers can take hundreds of lines, making it difficult for a human to use these to understand the underlying (mal)functioning of the system. 
There is some work along the lines of designing provers which produce proofs that look like ones humans might write~\cite{Horacek:2007,Ganesalingam:Gowers:2017}, but only for very specific domains. 

In terms of machine-to-machine explanations, one can consider the example of proof-carrying code, where an application downloaded from an untrustworthy location comes with a proof of its “correctness” for the host system to verify before installation. No human intervention is needed in order to either generate the proof or to verify it. This can be considered an example of the general idea of an interactive proof system~\cite{Goldwasser.et.al:2019} as follows. 
An interactive proof system models computation as the exchange of messages between two parties: a prover and a verifier. 
The prover has vast computational resources, but cannot be trusted, while the trusted verifier has bounded computation power. As the name suggests, the prover tries to prove some statement to the verifier. Interactive proofs often proceed in “rounds”, where the parties send messages to each other based on previous messages they have received, till the verifier is “convinced” of the truth of the statement. One can also have one-round (often referred to as non-interactive) proofs where the prover only needs to send one message to convince the verifier of a true statement, and no further rounds of interaction are necessary. 

Interactive proof systems have two requirements: soundness and completeness. Essentially, soundness claims that no prover, even a dishonest one, can convince the verifier of a false statement. Completeness says that for every true statement, there is a proof that the prover can produce to the verifier to convince it. 
The verifier may choose to “probe” various parts of the proof sent by the prover to convince itself of the verity of the statement. This is often done in an efficient manner by picking random bits and using them to identify which parts of the proof to inspect. Depending on what the abilities of the prover and the verifier are, one can get different classes of proofs. One can get (slightly) different systems based on whether the random values chosen by the verifier are made public or kept private. If one assumes the existence of special objects like one-way functions, one can construct “zero-knowledge proofs”, where the verifier is convinced exactly of the intended statement, but no further information about said statement is revealed. One can also have systems where multiple provers can interact with the verifier (but not with each other) to prove a statement. 
If proofs as above are considered as explanations, then we can categorize them as \textbf{attributive} explanations. 
Interactive proof systems can thus be seen as an excellent example of machine-to-machine explainability approaches that have existed for a long time.

\subsection{Argumentation}
\label{subsec:Argumentation}

“Computational Argumentation is a logical model of reasoning that has its origins in philosophy and provides a means for organizing evidence for (or against) particular claims (or decisions).”\cite[p.~277]{Sklar:Azhar:2018} 
\citeauthor{Moulin.et.al:2002} review in~\cite{Moulin.et.al:2002} a body of literature on explanations in knowledge based-systems and advocate the use of argumentation to support interactive and collaborative explanations of reasoning that take into account the aspects of justification as well as criticism of claims/decisions/solutions. 
Indeed, “[t]here is a natural pairing between Explainable AI and Argumentation: the first requires the need to explain decisions and the second provides a method for linking any decision to the evidence supporting it.”\cite[p.~277]{Sklar:Azhar:2018} 

By and large, data and knowledge in argumentation formalisms can be represented using various forms of directed graphs, whereby nodes called \emph{arguments} represent individual pieces of information and edges represent relationships among arguments (e.g.\ \emph{attacks} for conflicting information, \emph{supports} for supporting information). 
Reasoning then amounts to finding sets of acceptable arguments, or ranking arguments in terms of their acceptability. 
Argument acceptability is defined according to argumentation semantics which comprise  collections of formal criteria that a set of acceptable arguments has to satisfy (such as arguments within the set not attacking each other and defending against all attackers outside the set; see e.g.~\cite{Baroni:Giacomin:2007,Baroni:Rago:Toni:2019} for overviews of argumentation semantics). 

\begin{figure}
\centering
\begin{tikzpicture}
    \node at (1, 3) {$\graph$};
    
	\node at (2, 3) {$a$}; 
	\draw [fill = green, fill opacity = 0.3] (2, 3) ellipse (0.3cm and 0.3cm);
	\node at (2, 2) {$b$}; 
	\draw [fill = red, fill opacity = 0.3] (2, 2) ellipse (0.3cm and 0.3cm);
	\node at (1, 1.2) {$c$}; 
	\draw [fill = green, fill opacity = 0.3] (1, 1.2) ellipse (0.3cm and 0.3cm);
	\node at (3, 1.2) {$d$}; 
	\draw [fill = red, fill opacity = 0.3] (3, 1.2) ellipse (0.3cm and 0.3cm);
	\node at (3, 0.2) {$e$}; 
	\draw [fill = green, fill opacity = 0.3] (3, 0.2) ellipse (0.3cm and 0.3cm);
	
	\draw[-latex, thick] (2, 2.3) to (2, 2.7); 
	\draw[-latex, thick] (1.2, 1.4) to (1.8, 1.8); 
	\draw[-latex, thick] (2.8, 1.4) to (2.2, 1.8); 
	\draw[latex-latex, thick] (1.3, 1.2) to (2.7, 1.2); 
	\draw[-latex, thick] (3, 0.5) to (3, 0.9); 
	
	\node at (1, 0) {(i)}; 
	
	\node at (5, 3) {$\tree$};
	
	\node at (6, 3) {$P: a$}; 
	\draw[-] (6, 2.8) to (6, 2.5);
	\node at (6, 2.3) {$O: b$}; 
	\draw[-] (6, 2.1) to (6, 1.8);
	\node at (6, 1.6) {$P: c$}; 
	\draw[-] (6, 1.4) to (6, 1.1);
	\node at (6, 0.9) {$O: d$}; 
	\draw[-] (6, 0.7) to (6, 0.4);
	\node at (6, 0.2) {$P: e$}; 
	
	\node at (5, 0) {(ii)}; 
\end{tikzpicture}
\caption{
(i) An argument graph with nodes as arguments, labelled $a, b, c, d, e$, and directed edges as attacks. 
Slightly abusing the notation: $a$ is defended from $b$ by $c$; $c$ also defends itself against $d$; $e$ defends $c$ from $d$. 
Arguments accepted (resp.\ rejected) under the grounded semantics are colored green (resp.\ red): $G = \{ a, c, e \}$ is the grounded extension of $\graph$.  
(ii) A grounded dispute tree between the proponent $P$ and opponent $O$, for the topic argument $a$. 
The proponent can successfully defend the claimed argument $a$ against all the counterarguments by using the arguments from $G$. 
}
\label{fig:arg}
\end{figure}
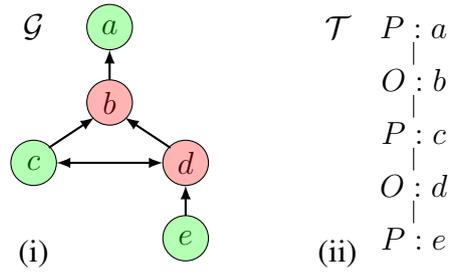

For concreteness, we consider the following setting. 
Let $\graph = \tuple{\Args, \attacks}$ be a directed graph representing an abstract~\cite{Dung:1995} or structured (see~\cite[Part II]{Rahwan:Simari:2009},~\cite{Besnard.et.al:2014}) argumentation framework with a set $\Args$ of arguments and an attack relationship $\attacks \subseteq \Args \times \Args$. 
    %
    %
    %
    %
We say that a set $S \subseteq \Args$ of arguments \emph{defends} an argument 
$x \in \Args$ just in case $\forall y \in \Args$ such that $y \attacks x$ there is $z \in S$ with $z \attacks y$. 
We assume that the acceptance of arguments in $\graph$ is evaluated using some \emph{extension}-based semantics, for instance the grounded semantics~\cite{Dung:1995}: 
    %
    %
the grounded extension of $\graph$ can be constructed as $G = \cup_{i \geqslant 0} G_i$, 
where $G_0$ is the set of all unattacked arguments, and $\forall i \geqslant 0$ $G_{i+1}$ is the set of arguments that $G_i$ defends; 
see Figure~\ref{fig:arg}(i). 
For any $\graph$, the grounded extension $G$ always exists and is unique. 

\subsubsection{Attributive/Contrastive argumentative explanations}
\label{subsubsec:arg contrastive}

A common approach to explaining argument acceptance in argumentation, one which encompasses both justification and criticism, essentially amounts to traversing (a part of) an argument graph to show the attacking and defending arguments (together with their relationships) relevant to accepting or rejecting an argument. 
Intuitively, to explain acceptance of an argument $x$, “one would need to show how to defend $x$ by showing that for every argument $y$ that is put forward (moved) as an attacker of $x$, one must move an argument $z$ that attacks $y$, and then subsequently show how any such $z$ can be reinstated against attacks (in the same way that $z$ reinstates $x$). 
The arguments moved can thus be organised into a graph of attacking arguments that constitutes an explanation as to why $x$ is [acceptable]”\cite[p.~109]{Modgil:Caminada:2009}. 
Such explanations thus aim to answer the following question~\cite{Fan:Toni:2015,Fan:Toni:2015-TAFA}:
\begin{itemize}[series=arg questions]
    \item Given an argumentation framework $\graph$, what are the reasons for and against accepting a particular argument $x$?
\end{itemize}
The sub-graph to be traversed for explaining acceptability of $x$ would normally satisfy some formal property of argument graphs (such as being a tree rooted in $x$, alternating attacking and defending arguments, with leaves holding unattacked arguments). 

Formally, let $a \in \Args$ be a \emph{topic} argument whose acceptance status requires an explanation. 
Consider a property $\prop$ applicable to sub-graphs of $\graph$. 
Following~\cite{Cocarascu.et.al:2018}, an explanation for the acceptance status of $a$ in $\graph$ can be defined as
\begin{quote}
    a sub-graph $\graph' = \tuple{\Args', \attacks'}$ of $\graph$ such that $a \in \Args'$ and $\graph'$ satisfies $\prop$. 
\end{quote}
Concrete properties of argument graphs yield explanations of concrete forms. 
Perhaps the most prominent form is achieved via a \emph{dispute tree} \cite{Dung:Kowalski:Toni:2006,Dung:Mancarella:Toni:2007}, defined for $a \in \Args$ as an in-tree\footnote{A directed rooted tree with edges oriented towards the root.} $\tree$ such that:
\begin{enumerate}[1.]
    \item Every node of $\tree$ is labelled by an argument and assigned the status of, exclusively, either proponent $P$ or opponent $O$ node;
    \item The root of $\tree$ is a $P$ node labelled by $a$, denoted $[P:a]$;
    \item For every node $[P:x]$ and for every $y \in \Args$ such that $y \attacks x$, the node $[O:y]$ is a child of $[P:x]$;
    \item Every node $[O:y]$ has exactly one child $[P:z]$ with $z \in \Args$ such that $z \attacks y$;
    \item There are no other nodes in $\tree$ except those given by 1-4 above.
\end{enumerate}
Notably, dispute trees often carry theoretical guarantees pertaining to desirable properties of explanations~\cite{Modgil:Caminada:2009,Fan:Toni:2015-TAFA,Cyras:Satoh:Toni:2016-COMMA,Cyras.et.al:2019-ESWA}, 
such as existence, correctness (the explanation actually establishes the acceptance status of the topic argument) and relevance (all the arguments in the explanation are relevant) among others. 

Depending on the acceptability status of the topic argument $a$, variants of dispute trees can be constructed~\cite{Dung:Kowalski:Toni:2009,Modgil:Caminada:2009} and used to define different forms of explanations~\cite{Dung:Mancarella:Toni:2007,Dung:Kowalski:Toni:2009,Modgil:Caminada:2009,Cyras:Satoh:Toni:2016-KR}. 
For example, if $a$ is in the grounded extension $G$ of $\graph$, there is a \emph{grounded} (i.e.\ finite~\cite{Dung:Kowalski:Toni:2009}) dispute tree $\tree$ for $a$; see Figure \ref{fig:arg}(ii). 
A grounded dispute tree $\tree$ contains the reasons sufficient to establish the acceptance of the topic argument: these are the proponent's $P$ arguments and they belong to the grounded extension $G$ of $\graph$~\cite{Dung:Kowalski:Toni:2009}.   
Thus, on the one hand, a dispute tree $\tree$ for $a$ can be seen as an \textbf{attributive} explanation for the acceptability of $a \in \Args$ in $\graph = \AF$.

On the other hand, dispute trees may contrast different acceptability statuses of the topic argument. 
For instance, a maximal~\cite{Cyras:Satoh:Toni:2016-KR} dispute tree explains why $a \in \Args$ is not accepted under grounded semantics in an acyclic $\graph$ and indicates which arguments need to be countered to make $a$ accepted. 
Alternatively, under other argumentation semantics, $\graph$ may have multiple extensions (i.e.\ sets of acceptable arguments), whence a particular argument may belong to one but not another. 
There would then be explanations for such alternative choices of acceptability~\cite{Modgil:Caminada:2009}: 
for example, if $\graph$ in Figure~\ref{fig:arg}(i) did not contain argument $e$, then $\graph$ would have two preferred~\cite{Dung:1995} extensions $\{ c, a \}$ and $\{ d, a \}$, and different dispute trees would explain e.g.\ the (non-)acceptance of $c$. 
Thus, a dispute tree $\tree$ for $a$ can also be seen as a \textbf{contrastive} explanation for the acceptability status of $a \in \Args$ in $\graph = \AF$.

More generally, dispute trees are a form of  \emph{dialogical} argumentative explanations (also called dialogue games or trees, \emph{dialectical} explanations or trees)~\cite{Prakken:Sartor:1998,McBurney:Parsons:2002,Walton:2004,Modgil:Caminada:2009,Garcia.et.al:2013,Booth.et.al:2014,Cyras:Satoh:Toni:2016-COMMA,Sklar:Azhar:2018,Sakama:2018,Cocarascu:Cyras:Toni:2018,Cyras.et.al:2019-ESWA,Zhong.et.al:2019,Cocarascu:Rago:Toni:2019,Madumal.et.al:2019,Cocarascu.et.al:2020}.
Dialogical explanations are argument exchanges between fictitious participants seeking to establish or explain the acceptability statuses of arguments. 
For example, the dispute tree $\tree$ from Figure~\ref{fig:arg}(ii) can be read as a dialogical explanation thus: 
``Proponent $P$ claims $a$; Opponent $O$ moves the only attacker $b$; $P$ responds with $c$, which $O$ counters with the only attacker $d$; $P$ responds with $e$ to which $O$ has no counter-arguments; therefore, the unobjectionable $e$ is accepted, and so are $c$ and $a$, whence $P$ wins". 

Dialogical explanations may, but need not be based on dispute trees. 
They can be used to explain both accepted and rejected arguments (e.g.~\cite{Modgil:Caminada:2009,Garcia.et.al:2013,Cyras:Satoh:Toni:2016-KR,Cyras:Satoh:Toni:2016-COMMA,Cyras.et.al:2019-ESWA}) 
and are applicable to other forms of argumentation, 
such as bipolar argumentation frameworks~\cite{Amgoud:Cayrol:Lagasquie-Schiex:2004,Cayrol:Lagasquie-Schiex:2009,Cohen.et.al:2018} that incorporate a support relationship in addition to attacks (e.g.~\cite{Cocarascu.et.al:2018,Rago:Cocarascu:Toni:2018}) 
as well as gradual argumentation~\cite{Amgoud:Ben-Naim:2018,Baroni:Rago:Toni:2019,Potyka:2019} that quantifies acceptability of arguments (e.g.~\cite{Cocarascu:Rago:Toni:2019}). 

Other versions of contrastive argumentative explanations can take forms of (possibly bounded) branches, paths or cycles in $\graph$ that present the most relevant information in favour of or against accepting a particular argument/claim~\cite{Seselja:Strasser:2013,Timmer.et.al:2017,Cocarascu.et.al:2018}. 
This role can also be assumed by sets of arguments (or attacks) themselves, keeping the attack or defence relationships (or relevant arguments) implicit~\cite{Fan:Toni:2015-TAFA,Zeng.et.al:2018-AAMAS,Cyras.et.al:2019-AAAI,Raymond:Gunes:Prorok:2020}. 

We finally remark on another kind of argumentative explanations. 
Recall that the argument graph $\graph$ considered above can represent an abstract or a structured argumentation framework. 
In the former, the arguments are atomic entities, as in Figure~\ref{fig:arg}. 
By contrast, in structured argumentation (e.g.~\cite[Part II]{Rahwan:Simari:2009},~\cite{Besnard.et.al:2014}), arguments have internal structure and relationships among arguments are constructively defined. 
Typically, an argument would comprise a set of premises, a conclusion and some form of linking the conclusion to the premises~\cite[pp.~183-193]{Moulin.et.al:2002}. 
These can be formalised in some formal logic, whence an argument is a sequence of inferences or a (defeasible) proof,
or by using argument instantiation mechanisms such as argument schemes~\cite{Walton:Reed:Macagno:2008}. 
Seen this way, the structure of an argument can itself provide an attributive explanation as justification or trace~\cite{Moulin.et.al:2002}. 
At the same time, however, the arguments' components and (logic- or argument scheme-based) relationships thereof can serve to construct dialogical explanations at the level of argument structure~\cite{Cyras.et.al:2018-HOFA,Collins:Magazzeni:Parsons:2019,Sassoon.et.al:2019,Kokciyan.et.al:2020}.

\subsubsection{Actionable argumentative explanations}
\label{subsubsec:arg actionable}

Another form of argumentative explanations allows one to speculate as to the changes to the argument graph, such as additions or removals of arguments or relationships, that result in different acceptance status(es) of the argument(s) in question. 
Such \textbf{actionable} argumentative explanations thus aim to answer the following question~\cite{Fan:Toni:2015-TAFA,Sakama:2018}:
\begin{itemize}[resume=arg questions]
    \item Which arguments/relationships should be removed or added to accept argument $a$?
\end{itemize}

Concretely, following~\cite{Fan:Toni:2015-TAFA}, let $a \in \Args$ be a topic argument whose acceptance status is undesirable. 
Specifically suppose $a$ is rejected in $\graph$ under the given semantics and that an explanation is wanted as to how to change the status, i.e.\ make $a$ accepted. 
An explanation can then be defined in several ways, for instance as:
\begin{quote}
    a set $\Args' \subseteq \Args$ of arguments such that $a$ is accepted in the
    sub-graph $\graph' = \tuple{\Args \setminus \Args', \attacks'}$ of $\graph$
    (where $\attacks'$ is a restriction of $\attacks$ over $\Args \setminus \Args'$); 
\end{quote}
\begin{quote}
    a set of attacks $\attacks' \subseteq \attacks$ such that $a$ is accepted in the
    sub-graph $\graph' = \tuple{\Args, \attacks \setminus \attacks'}$ of $\graph$. 
\end{quote}
For example, intuitively, 
the attack $c \attacks b$ in Figure~\ref{fig:arg}(i) explains why $b$ is rejected (under grounded semantics) in $\graph$, but would be accepted if the attack were removed. 

More generally, explanations can exist not only for how to make an argument accepted, but also rejected or having yet other acceptance status~\cite{Wakaki:Nitta:Sawamura:2009,Sakama:2018,Saribatur.et.al:2020}. 
For example, in Figure~\ref{fig:arg}(i), argument $c$ explains why $a$ is accepted, but would be rejected if $c$ (and the adjecent attacks) were removed. 
Such explanations can also make use not only of removal, but of addition of arguments and/or attacks~\cite{Wakaki:Nitta:Sawamura:2009,Booth.et.al:2014,Sakama:2018}, for instance assuming a pre-specified space from which arguments/attacks can be added. 
Further, various notions of minimality of removal and/or addition can be imposed~\cite{Fan:Toni:2015-TAFA,Sakama:2018} to make explanations more actionable. 


Actionable argumentative explanations are clearly related to dialogical explanations in that, intuitively, the actionable explanations indicate how to obtain a ``good" dispute tree~\cite{Fan:Toni:2015-TAFA} or a ``good" move in a dialogue game~\cite{Sakama:2018}. 
Such explanations are also very closely related to argumentation dynamics~\cite{Rotstein.et.al:2008,Cayrol:Saint-Cyr:Lagasquie-Schiex:2010,Baroni.et.al:2014} and the so-called enforcement problems~\cite{Baumann:Brewka:2010,Baumann:2012,Coste-Marquis.et.al:2015}, which deal with changes to argumentation frameworks that change the statuses of arguments under various semantics~\cite{Ulbricht:Baumann:2019}. 
Although explainability as such is not considered in these works, some formal relationships have been established in e.g.~\cite{Sakama:2018}. 
This seems to be a promising area of research on actionability of argumentative explanations.

\subsubsection{Applications of argumentative explanations}
\label{subsubsec:arg applications}

There is a wealth of studies and applications of argumentative explanations in modern MR. 
Alongside the works supplying theoretical foundations cited above, explanations in and for argumentation itself have been studied in e.g.\ the following works:~\cite{Moulin.et.al:2002,Walton:2004,Walton:2009,Modgil:Caminada:2009,Wakaki:Nitta:Sawamura:2009,Garcia.et.al:2013,Booth.et.al:2014,Garcia:Simari:2014,Fan:Toni:2015,Fan:Toni:2015-TAFA,Cyras:Satoh:Toni:2016-KR,Cyras:Satoh:Toni:2016-COMMA,Bex:Walton:2016,Sakama:2018,Zeng.et.al:2018-AAAI,Karamlou:Cyras:Toni:2019-demo,Sassoon.et.al:2019,Morveli-Espinoza:Tacla:2020,Kokciyan.et.al:2020,Liao:Torre:2020,Saribatur.et.al:2020}. 
Applications of argumentative explanations to other areas of AI, including but not limited to MR, as well as adjacent fields are numerous too.
The following are examples of settings and works where argumentative explanations have been applied to explain reasoning: 
\begin{enumerate*}[1.]
    \item Forms of \nameref{subsec:LP}~\cite{Booth.et.al:2014}, particularly \nameref{subsubsec:ASP}~\cite{Wakaki:Nitta:Sawamura:2009,Schulz:Toni:2013,Schulz:Toni:2016}, 
    Defeasible LP~\cite{Garcia:Simari:2014}; 
    
    \item database query answering~\cite{Arioua:Tamani:Croitoru:2015,Hecham:2017}; 
    
    \item forms of decision making~\cite{Amgoud:Prade:2009,Zeng.et.al:2018-AAMAS,Zhong.et.al:2019}; 
    
    \item AI planning~\cite{Fan:2018,Collins:Magazzeni:Parsons:2019} and scheduling~\cite{Cyras.et.al:2019-AAAI,Cyras.et.al:2020-AAMAS}; 
    
    \item explainable agents~\cite{Sklar:Azhar:2018,Madumal.et.al:2019,Raymond:Gunes:Prorok:2020};
    
    \item classification tasks~\cite{Amgoud:Serrurier:2008,Cocarascu:Cyras:Toni:2018,Sendi.et.al:2019,Cocarascu.et.al:2020,Prakken:2020,Albini.et.al:2020};
    
    \item recommendation~\cite{Briguez.et.al:2014,Rago:Cocarascu:Toni:2018,Cocarascu:Rago:Toni:2019} and link prediction~\cite{Albini.et.al:2020-XLoKR} tasks; 
    
    \item legal informatics~\cite{Timmer.et.al:2017,Cyras.et.al:2019-ESWA,Collenette:Atkinson:Bench-Capon:2020};
    
    \item medical reasoning~\cite{Sassoon.et.al:2019,Kokciyan.et.al:2020};

    \item scientific debates~\cite{Seselja:Strasser:2013}.
\end{enumerate*}
By and large, attributive and contrastive (mostly dialogical) explanations prevail in argumentation and its applications. 
We instead call for research in and applications of actionable argumentative explanations.

\subsection{Planning}
\label{subsec:Planning}

Automated planning (or AI planning) is a class of decision making techniques that deals with computing sequences of actions towards achieving a goal. 
The solution to a planning problem is found by a planner, typically one based on heuristic search. 
Explainability of an AI planning system \AI\ concerns the workings of the planning algorithm, details of the underlying model of the planning problem as well as the reconciliation of the differences between the system’s model and the user's (mental) model~\cite{Chakraborti.et.al:2020}. 
The kinds of questions that explanations in planning aim to answer pertain to unsolvability, alternative choices and the goodness of plans, as follows:
\begin{itemize}
    \item Why did the agent take action $A$ (at that time) rather than action $B$ (resp.\ earlier or later)?~\cite{Cashmore.et.al:2019}
    \item What are possible reasons the planner could not compute a plan from the current state to the given goal state?~\cite{Gobelbecker.et.al:2010}
    \item Why does a plan fail to be a solution, which actions are invalid?~\cite{Fan:2018}
    \item Why fail and what (temporal) repercussions does the first failure have?~\cite{Torta.et.al:2019}
    \item What changes (in the current state) would make the planning problem solvable?~\cite{Gobelbecker.et.al:2010}
    \item How good is a given plan from the point of view of the human observer (their computational model)?~\cite{Chakraborti.et.al:2019}
\end{itemize}

In~\cite{Chakraborti.et.al:2020}, \citeauthor{Chakraborti.et.al:2020} provide an excellent overview of the most recent advances in explainable AI planning (XAIP). 
They emphasize the importance of user modelling, or “persona” of the explainee. Although their classification is broad -- in terms of end-user, algorithm designer and model designer personas -- it is clear that more granular models are possible. 
Their work reinforces~\citeauthor{Miller:2019}'s view on the characteristics of effective explanations~\cite{Miller:2019}, namely that explanations should be ``social in being able to model the expectations of the explainee, selective in being able to select explanations among several competing hypothesis, and contrastive in being able to differentiate properties of two competing hypothesis.''\cite[p.~4805]{Chakraborti.et.al:2020} 
Further, \citeauthor{Chakraborti.et.al:2020} also point to the use of abstractions as a means to provide effective explanations. 
We highly recommend the reader to consult~\cite{Chakraborti.et.al:2020} for approaches to XAIP, indifferently ML- and MR-based. 
Instead, we here show how explanations in planning fall into the categories of explanations that we proposed, and briefly exemplify with some MR-based XAIP approaches. 

Following~\cite{Chakraborti.et.al:2020}, we assume a simplified, abstract representation of a planning problem and a planning system \AI\ (planning agent or planner) thus. 
A planning problem $\Pi$ can be implicitly defined via a transition function $\delta: S \times A \to S$, with $S$ a set of states and $A$ a set of actions available to the agent. 
Solving $\Pi$ amounts to producing a plan $\pi = \tuple{a_0, \ldots, a_n}$ as a sequence of actions $a_k \in A$ that allows transitioning from the initial state $I \in S$ to a goal state $G \in S$, subject to some desired properties such as optimality. 
(Alternatively, $\pi$ can be a policy mapping states to actions that the agent should take.)
We may assume that the initial and goal states, together with properties to optimize, can be represented by information $i$, whence the AI planning system \AI\ produces outcome $o = \pi$ given $i$. 

To answer why \AI\ yields a plan $\pi$ given $i$, a (positive) \textbf{attributive} explanation can be a path induced by $\pi$ in the state space from the initial state $I$ to the goal state $G$, and a proof that the path satisfies the desired properties.
When no plan is possible to achieve the goal state, a (negative) \textbf{attributive} explanation can be a property which is true of all the paths starting from the initial state and a proof that the goal is unreachable in any path with this property. 
For example, there may not be any action that achieves a specific predicate which is part of the goal state, or the predicates in the goal state are rendered mutually-exclusive by every action. 

Alternatively, to answer questions about unsolvability of the planning problem $\Pi$, some approaches provide \textbf{contrastive} explanations in terms of a part $\Pi'$ of the problem model $\Pi$ that together with the initial state $i$ and the desired goal(s) lead to unsolvability. 
For instance, \citeauthor{Fan:2018} in~\cite{Fan:2018} captures STRIP-like planning in Assumption-Based Argumentation and extracts explanations pertaining to actions and/or their preconditions and goals relevant to both validity and invalidity of a plan. 
The unsolvability of planning tasks can also be explained to the user by pointing out unreachable but necessary goals in terms of propositional fluents~\cite{Sreedharan.et.al:2019}. 
Similarly, pointing out actions executed in a faulty node and their propagations (subsequent failed actions and their relationships) can be used to define explanations in temporal multi-agent planning~\cite{Torta.et.al:2019}. 
Such explanations mostly work by attributing counterexamples to solvability in the planning system \AI.

However, counterfactual \textbf{contrastive} explanations are also possible. 
They can for instance work in terms of \emph{excuses}---minimal, restrictive changes to the initial state $i$ that would allow for a plan $\pi$ that reaches the desired goals~\cite{Gobelbecker.et.al:2010}.
More generally, contrastive explanations can work via alternative courses of action and properties satisfied or not by the (current or alternative) plan. 
The basic idea in answering questions about a given plan $\pi$ not satisfying some property $P$ is to produce an alternative plan $\pi'$ that does satisfy $P$ and compare the two plans by showing that either some goal is not reachable via $\pi'$ or that $\pi'$ is in some sense less optimal than $\pi$~\cite{Eifler.et.al:2020-AAAI}.
For instance, in~\cite{Eifler.et.al:2019}~\citeauthor{Eifler.et.al:2019} develop a means to qualify how good or poor a plan is, beyond the obvious properties such as cost or plan length. 
In the specific case of goal exclusions, explanations amount to $\subseteq$-minimal unsolvable goal subsets. 
Such plan-property dependencies allow oversubscribed goals to be reasoned over, and explained by an agent. 
Specifically, in~\cite{Eifler.et.al:2020-AAAI}~\citeauthor{Eifler.et.al:2020-AAAI} define explanations via plan-property entailment of soft goals so that an answer to the question ``why is a property (set of soft goals) not achieved?" amounts to exhibiting other properties (sets of soft goals) that would not be achieved otherwise. 
This can be extended to plan properties formulated in e.g.\ linear temporal logic~\cite{Eifler.et.al:2020-IJCAI}. 

Another prominent challenge in explaining planning that has recently attracted a fair amount of research interest is that of model reconciliation~\cite{Chakraborti.et.al:2017,Chakraborti.et.al:2019,Kulkarni.et.al:2019,Chakraborti.et.al:2019-AIComms,Chakrabort.et.al:2019-HRI}. 
It aims to address the questions pertaining to the goodness of a given plan with respect to the user's computational model, typically human's mental model. 
Here the assumption is that humans have a domain and task model that differs significantly from that used by \AI, and explanations suggest changes to the human’s model to reconcile the differences between the two. 
The objective of model reconciliation is not to completely balance the information asymmetry, but is selective to knowledge updates that can minimally cause human-computed plans to match those computed by the system. 
Such explanations are \textbf{contrastive} too~\cite{Chakraborti.et.al:2020}, in that they contrast the models of the AI system and its user and aim to bring modification to the latter so as to convince the user of the goodness of the plan. 
Model reconciliation has also been addressed as a~\nameref{subsec:CP} explainability problem using \nameref{subsubsec:ASP}~\cite{Nguyen.et.al:2020-KR,Nguyen.et.al:2020-AAMAS}.

We note that model reconciliation is in some aspects closely related to plan/goal recognition~\cite{Carberry:2001}. 
Essentially, recognizing the AI agent's goals may be seen as a prerequisite to providing rationale for its behavior. 
And the other way round, recognizing the user's plans and goals can serve as a means to improve explanations. 
For instance, \citeauthor{Krarup.et.al:2019} show in~\cite{Krarup.et.al:2019} how user constraints can instead be added to the formal planning model so that contrastive explanations as differences between the solutions to the initial and the new model can be extracted. 
So while model reconciliation aims at changing the user's model, the other direction of changing the AI system's model could lead to actionable explanations. 
For example, when no plan is possible for a goal state and there is an explanation for the unsolvability, an actionable explanation could pertain to modification of the action space so that a plan could be found in the new model of the problem. 
Overall, however, to our understanding there do not seem to be actionable explanations in AI planning yet.

\subsection{Decision Theory}
\label{subsec:DT}
\label{subsec:CSC}

Decision theory gathers different domains such as Multi-Criteria Decision Making, Decision Making under Uncertainty and Computational Social Choice (SC). 
``The typical decision problem studied in decision theory consists in selecting one alternative among a set $X$ of candidate options, where the alternatives are described by several dimensions. This selection is obtained by the construction of a preference relation over $X$.''\citep[p.~1411]{Labreuche:2011} 
The final part of the decision process, i.e.\ explaining the outcome of the decision model to the user, is by and large not readily supported by decision theory models due to their complexity. 
It is nonetheless arguably as important as the recommendation of the outcome itself~\cite[p.~152]{Belahcene.et.al:2017} and attempts at answering the following question:
\begin{itemize}
    \item Can this recommendation (i.e.\ the chosen alternative) be justified under the given preference profile?~\cite{Labreuche:2011,Boixel:Endriss:2020}
\end{itemize}

Explainability in decision theory has been researched in modern MR. 
On the one hand, forms of argumentation have been proposed for explainable decision making, see e.g.~\cite{Amgoud:Serrurier:2008,Amgoud:Prade:2009,Zhong:Fan:Toni:Luo:2014,Zhong.et.al:2019}. 
Specifically, in~\cite{Amgoud:Serrurier:2008,Amgoud:Prade:2009} the authors propose to use abstract argumentation with preferences for multiple agents to argue about acceptable decisions. 
There, explainability is assumed to arise naturally from the argumentative process, but is not specified at all. 
Instead, \citeauthor{Zhong.et.al:2019} map decision frameworks into Assumption-Based Argumentation and use dispute trees (see Section~\ref{subsec:Argumentation}) to generate explanations as to why a decision is best, better than or as good as  others~\cite{Zhong.et.al:2019}. 
They also translate such \textbf{contrastive} dialogical explanations into natural language and perform an empirical user study in a legal reasoning setting to evaluate their approach. 
These works effectively define argumentation-supported decision making procedures where explainability arises from the argumentative methods employed. 

On the other hand, \citeauthor{Labreuche:2011} in~\cite{Labreuche:2011} provides solid theoretical foundations to produce explanations for a range of decision-theoretic models via provision of arguments, but not using formal argumentation. 
There, arguments, and therefore explanations, essentially are ``based on the identification of the decisive criteria."\citep[p.~1415]{Labreuche:2011} 
They are thus \textbf{attributive}, supplying justifications in terms of higher-level criteria that support the recommended decision. 
Similar in spirit, \citeauthor{Labreuche:Maudet:Ouerdane:2011} in~\cite{Labreuche:Maudet:Ouerdane:2011} consider the setting of qualitative multi-criteria decision making with preferential information regarding the importance of the criteria and preference rankings over different options (choices). 
They define explanations as collections of factored preference statements (roughly of the form `on criteria $I$, option $o$ is better than options $P$') that justify the choice of the weighted Condorcet winner. 
Such explanations pertain to the relevant data (problem inputs) supporting a proof that the recommended decision/choice is the best one. 

Some more recent works essentially apply to various settings the same conceptual idea that explanations of decisions made pertain to important criteria or principles based on which the model makes a decision. 
We deem such explanations pertaining to be \textbf{attributive}. 
For instance, in~\cite{Nunes.et.al:2014} explaining amounts to identification of decisive criteria as sets of attributes that are most important for preferring one option over another. 
\citeauthor{Nunes.et.al:2014} thus propose attributive explanations via several decision criteria in a quantitative multi-attribute decision making setting.  
\citeauthor{Belahcene.et.al:2017} deal with incomplete preference specifications in~\cite{Belahcene.et.al:2017} and thereby use preference swaps to explain/justify decisions. 
Intuitively, their explanations transform a complex preference statement (over many attributes) that needs to be understood by the user into a series of simpler preference statements (over few attributes). 
In~\cite{Labreuche:Fossier:2018}, \citeauthor{Labreuche:Fossier:2018} consider hierarchical models of multi-criteria decision aiding. 
They use Shapley values to define axiomatic indices regarding the influence of different criteria and provide attributive explanations pertaining to the importance of different criteria. 
In a setting where different audiences may adhere to different norms~\cite{Boixel:Endriss:2020}, \citeauthor{Boixel:Endriss:2020} explain the decision making outcome by presenting axioms (with respect to audience's norms) for which no voting rule would yield a different outcome.

\subsection{Causal Approaches}
\label{subsec:Causal}

Causal models (see e.g.~\cite{Geffner:1990,Lacave:Diez:2002,Pearl:2019}) are useful in guiding interventional decisions and analyzing counterfactual hypothetical situations. 
Using causal models one can not only provide a decision but also provide a basis for what-if analysis, thus providing explanations. 
\citeauthor{Lacave:Diez:2002} provide in~\cite{Lacave:Diez:2002} a summary of the work done on explaining AI models where the models are causal, quite often Bayesian networks. 
The authors distinguish three classes of explanations.
\begin{enumerate*}
    \item Explanation of evidence -- this is basically abduction where the explanation is finding most probable explanations of variables not directly observable, based on the observed evidence variables. 
    This can be seen as a form of \textbf{attributive} explanations. 
    \item Explanation of the model -- this is simply a description (graphically or in text) of the causal model.
    \item Explanation of reasoning -- here the objective is to explain the reasoning process for the result obtained, for specific results not obtained, or for counterfactual reasoning. 
    We see these largely as \textbf{contrastive} explanations. 
\end{enumerate*}

As regards the first class, \cite{Geffner:1990} can be seen as an early example. 
There, \citeauthor{Geffner:1990} equates “being caused” with “being explained”. 
In contrast to earlier explanation techniques that used  logical derivations (roughly as antecedents of rules that explain the consequents), \citeauthor{Geffner:1990} augments default theories~\cite{Reiter:1980} with a causality/explanations operator $C$ which is used to define an order over classes of models of default theories in terms of explained abnormalities ($ab$ predicates). 
These kind of rule-based attributive explanations therefore have an abductive flavor.

A modern MR example to explaining causality is~\cite{Nielsen.et.al:2008}. There \citeauthor{Nielsen.et.al:2008} show that given a set of variables, the values of which an explanation is sought after, it is possible to determine a set of other variables (explanatory variables) which probabilistically explain the given set of observed variables. 
Some of these are possibly observed while the others are not. 
Indeed this includes abducing some variables from others. The use of a causal Bayesian network to trace these other variables is particularly interesting in the case of causal explanations. 
\citeauthor{Nielsen.et.al:2008} further show how to use the interventional distribution on a Causal Bayesian network to compute a causal explanation, thus approaching the realm of contrastive explanations. 

Another recent example of attributive explanations with a counterfactual flavor is the work of~\cite{Madumal.et.al:2020} which uses structural causal models to explain actions of Reinforcement Learning agents. 
In principle, to support counterfactual reasoning and thus \textbf{contrastive} explanations, one can use pure regression techniques factoring in time to do some causal correlation. 
For instance, in Granger-causal inferencing the idea is to use time series data analysis and hypothesis testing to extract possible causal influences.

However, much more informative models are structural equation models~\cite{Pearl:2019} and causal Bayesian networks~\cite{Hitchcock:2018}. 
The structural equation model allows specification of equations that denote the effect of one variable on the other. That is most helpful in doing interventional analysis. The model also supports a logic with an algebra that allows counterfactual analysis. 
On the other hand, Bayesian networks allow for probabilistic relations between variables and thus also enable interventional and counterfactual analysis. 
The Structural Causal Model is possibly the most evolved and combines the benefits of both these models. 
The rich set of analytical tools that it comes with is described in~\cite{Pearl:2019}.

Generally, \citeauthor{Pearl:2019} argues in~\cite{Pearl:2019} that explainability and other obstacles “can be overcome using causal modeling tools, in particular, causal diagrams and their associated logic.” Methods for creating such causal models are therefore of great importance. 
The main challenge of these methods in practice, however, is learning the model from data.  
Very often expert input is used in conjunction with data to build the models. 
\citeauthor{Heckerman:1995} describes in~\cite{Heckerman:1995} how it is possible to build Causal Bayesian Networks from data, under some assumptions. 
Causal Bayesian Networks can be extended to Bayesian Decision Networks where the decision variables and the utility (optimization) variables are explicitly identified and used for decision making.  
See e.g.~\cite{Constantinou:Fenton:2018} for a concise introduction to the area.

Overall, we contend that despite the progress with techniques for causality, we are still far from formalizing and being able to explain other more nuanced interpretations of causality that humans are familiar with. 
This is exemplified by the works on ‘actual’ causality, e.g.~\cite{Denecker.et.al:2019}, which illustrates with simple examples the challenges or explainability using simple causal diagrams.

\subsection{Symbolic Reinforcement Learning}
\label{subsec:RL}

Reinforcement learning (RL) has recently become visible as a promising solution for dealing with the general problem of optimal decision and control of agents that interact with uncertain environments. Application areas range from telecommunication systems, traffic control, autonomous driving, robotics, economics and games. In the general setting, an RL agent is usually operating in an environment repetitively applying one of the possible available actions and receives a state observation and a reward as feedback. 
The goal of the RL framework is to maximize the overall utility over a time horizon. The choices of right actions are critical; while some actions exploit the existing knowledge, some actions explore to how to increase the collected reward in future, at the cost of performing a locally sub-optimal behavior. 

Explaining control decisions produced by RL algorithms is crucial~\cite{Agogino:Lee:Giannakopoulou:2019}, since the rationale is often obfuscated, and the outcome is difficult to trust for two reasons: 
\begin{enumerate*}
\item lack of coverage in the exploration, and 
\item generalizability of the learned policies. 
\end{enumerate*}
Explainability in RL is complicated due to its real-time nature, since control strategies develop over time, and are typically not evaluated over snapshots. 
Several techniques for explanations are proposed in~\cite{Agogino:Lee:Giannakopoulou:2019} such as Bayesian rule lists, function analysis, Grammar-based decision trees, sensitivity analysis combined with temporal modeling using long-short term memory networks, and explanation templates. 
Albeit such techniques are relevant as early attempts towards explaining RL decisions, we do not review them in this report since we do not think they fall within MR. 
Instead we next briefly discuss some approaches that use symbolic techniques for explainability in RL. 
We also refer the reader to additionally consult the recent brief overview~\cite{Puiutta:Veith:2020} of explainability in RL. 

Approaches in the literature that integrate the RL framework with symbolic techniques for the purpose of explainability have been investigated in e.g.~\cite{Fukuchi.et.al:2017,Juozapaitis.et.al:2019,Khan:Poupart:Black:2009}. 
In particular, \citeauthor{Fukuchi.et.al:2017} introduce a framework of instructions-based behavior explanation in order to explain the future actions of RL agents. In this way, the agent can reuse the instructions from the human which leads to faster convergence. 
In~\cite{Juozapaitis.et.al:2019}, the method of reward decomposition is proposed in order to explain the actions taken by an RL agent. The idea is to split the reward in semantically meaningful types such that such that the action of the RL agent can be compared with reference to trade-offs between the rewards. 
The study in~\cite{Khan:Poupart:Black:2009} deals with the general framework of explaining the policies over Markov Decision Processes (MDPs). Under the assumption that the MDP is factored, a subset a minimum set of explanations that justify the actions of MDP is proposed. 
We believe these type of justifying explanations are a form of \textbf{attributive} explanations. 


There is recent work on \textbf{contrastive} explanations in RL too. 
For instance, \citeauthor{Carrillo:Rosenblueth:2007} show in \cite{Carrillo:Rosenblueth:2007} how model checking enables explanations via counterexamples regarding satisfaction of logical properties by Kripke models, which are closely related to MDPs. 
On the other hand, in~\cite{Madumal.et.al:2020} \citeauthor{Madumal.et.al:2020} define causal and counterfactual explanations for MDP-based RL agents given their action influence models (which extend structural causal models~\cite{Pearl:2019} with actions). 
Specifically, the authors define a (complete) causal explanation for an action $A$ “as the complete causal chain from $A$ to any future reward that it can receive”, with a minimal such explanation omitting intermediate nodes in such a chain, leaving source and destination nodes only. 
Further, they define a (minimally complete) explanation as the difference between the actual causal chain for the taken action $A$, and the counterfactual causal chain for some other action $B$. 
They show experimentally with human users that their explanations are subjectively good enough and help the users to better understand the RL agent’s actions. However, the method requires a correct model of the world to be given upfront.

\subsubsection{Constrained RL}
\label{subsubsec:constrained}

State-of-the-art RL is associated with several challenges: guaranteed safety during exploration in the real world is one of them, and intricacy of reward construction is another. Many recent works introduced preliminary results on mitigating these challenges through formal methods~\cite{Alshiekh.et.al:2018,Jansen.et.al:2018,Isele.et.al:2018,Li:Ma:Belta:2018}. 
The most related ones to MR focus on shielding or constraining exploration in general or in various specific contexts, presenting objectives in the form of a linear temporal logic (LTL) formula~\cite{Alshiekh.et.al:2018, Jansen.et.al:2018, alex_PhD, DBLP:journals/arobots/NikouHD20, alex_automatica_2017}. Recent works also include the design of control policies using RL methods to find policies which guarantee the satisfaction of properties described in temporal logic~\cite{Isele.et.al:2018, Li:Ma:Belta:2018, Bouton.et.al:2019}.

\subsubsection{Multi-Agent RL (MARL)}
\label{subsubsec:MARL}

When it comes to Multi-Agent RL (MARL) frameworks, the main challenge to be addressed is the dependency between the action and rewards of different agents in the environment~\cite{Lauer:Riedmiller:2000,Zhang:Yang:Basar:2019}. 
It is natural when two or more agents are trying to optimize their local behavior over a horizon of time, and conflicts might occur with respect to the team or global behavior of the agents. In such a setting, different local optimal actions might lead to conflicting collaborative behavior. Thus, such scenarios render the collaborative reward design challenging, and new algorithms should be designed in order to address such problems. 
The MARL scenario imposes additional constraints and an efficient way to handle them is to use a symbolic framework by presenting the constraints in a more convenient ways, such as logical description. 
\citeauthor{Kazhdan:Shams:Lio:2020} in~\cite{Kazhdan:Shams:Lio:2020} provide such model extraction techniques that enhance explainability of MARL frameworks.

\section{Discussion}
\label{sec:Discussion}

We here discuss various limitations of our overview and categorization of approaches to explainability in MR. 
First, we touch on some inevitable omissions regarding MR-related approaches. 
We then consider some aspects that are related to our proposed categorization of explanations. 
And finally we remark on the terminology (not) employed in this report.

\subsection{Omissions}
\label{subsec:Omissions}

In this report, we did not cover the following research areas that could potentially be considered related to MR.
\begin{enumerate}
    \item Rule-based classification/prediction, e.g.~\cite{Yin:Han:2013}, which comprises of largely ML-focused approaches for associative/predictive rule generation and/or extraction. Overviews of such rule-based ML explainability can be found in e.g.~\cite{Andrews:Diederich:Tickle:1995,Guidotti.et.al:2019,Adadi:Berrada:2018,Arrieta.et.al:2020}.
    
    \item Neuro-symbolic computing and graph neural networks, see e.g.~\cite{Lamb.et.al:2020} for an overview, where the use of classical logics alongside ML techniques has recently been argued as a factor enabling explainability. 
    Other works use classical logic for approximation and human-readable representation of ML workings. 
    For instance, the authors in~\cite{Ciravegna.et.al:2020-AAAI,Ciravegna.et.al:2020-IJCAI} use neural networks to learn relationships between ML classifier outputs and interpret those relationships using Boolean variables that represent neuron activations as well as predicates that represent the classification outputs. These then yield human-readable first-order logic descriptions of the learned relationships. 
    Neuro-symbolic computing comprises heavily of ML-focused approaches that we deemed beyond the scope of this report on MR explainability. 
    
    \item Probabilistic reasoning, which is a field (of Mathematics as well as Computer Science) in itself, and  spans AI at large. We met several explanation-oriented approaches when discussing various MR branches, but we do not separately consider Bayesian Networks or Probabilistic Graphical Models among MR branches with a focus on explainability. 
    
    \item Game theory, which is a field of Mathematics that studies strategic interaction between rational decision makers. In AI, game theory provides foundations to reasoning with preferences, multi-agent systems (MAS) and mechanism design among others, possibly with the benefit of enabling explainability. 
    SHAP (Shapley additive explanations)~\cite{Lundberg:Lee:2017} is a well-known example of application of game theory to explainability in ML as well as MR~\cite{Labreuche:Fossier:2018}. 
    We do not think that game-theoretic approaches constitute a branch of MR, or at least not one with a particular focus to explainability. 
    We did, however, briefly discuss some related approaches to explainability from~\nameref{subsec:DT}. 
    
    \item Robotics, which is an interdisciplinary branch of Engineering and Computer Science. In terms of explainability, the cornerstone notion there is explainable agency~\cite{Langley.et.al:2017}, which amounts to an autonomous agent being able to do the following:
    \begin{enumerate}
        \item explain decisions made during plan generation;
        \item report which actions it executed;
        \item explain how actual events diverged from a plan and how it adapted in response;
        \item communicate its decisions and reasons.
    \end{enumerate}
    In~\cite{Anjomshoae.et.al:2019} \citeauthor{Anjomshoae.et.al:2019} review explainable agency approaches where goal-driven agents and robots purport to explain their actions to a human user. 
    We encountered some approaches to explainability of autonomous agents which belong to branches of MR such as~\nameref{subsec:Argumentation} and \nameref{subsec:Planning}. 
    Generally, however, explainable agency considers human-AI interaction aspects which we do not cover in this report. 
\end{enumerate}

We hope these omissions do not lessen our contribution of a conceptual overview of MR explainability.

\subsection{Categorization-related Aspects}
\label{subsec:Aspects}

We here briefly discuss a few aspects related to our categorization of explanations (Section~\ref{subsec:Categorization}) and its relationship to MR approaches overviewed in this report.

\paragraph{Causality}

We have discussed~\nameref{subsec:Causal} loosely as a branch of MR where explainability is investigated. 
Instead, \emph{causality} can be viewed as a property or even constitute a category of explanations~\cite{Johnson:Johnson:1993}. 
However, following~\cite{Miller:2019,Wang.et.al:2019,Sokol:Flach:2020}, we contend that explanations may, but need not in general be causal. 
It is for instance acknowledged that “it does not seem possible to develop a formal connection between counterfactuals and causality.”\cite[p.~69]{Ginsberg:1986}  
We thus maintain that the aspect of causality is orthogonal to our categorization of explanations.

\paragraph{User-centrism}

Similarly to causality, we recognize the property of explanations being \emph{user-centric}, see e.g.~\cite{Moore:Swartout:1990,Chakraborti.et.al:2017,Ribera:Lapedriza:2019,Wang.et.al:2019,Zhou:Danks:2020,Mohseni:Zarei:Ragan:2020}. 
While it clearly applies to attributive, contrastive and actionable explanations, here we also have in mind a slightly different, more general notion of user-centrism. Specifically, it pertains to approaches to explainability directly taking into account the user (human or AI, indifferently) model, preferences, intentions etc. 
It is not only about the explanations being e.g.\ accessible, comprehensible, informative, but also about the relation between the explainee and the explanation. For instance, in the human user case, 
\begin{enumerate}
    \item an attributive explanation may need to take into account the relevance of the attributions to the user, e.g. domain vs procedural knowledge in the inference from knowledge to outcome; 
    \item a contrastive explanation may need to take into account the context in terms of which counterfactual situations are attainable to the user, e.g. the controllable aspects of the problem domain such as resource availability;
    \item an actionable explanation may need to take into account the user’s preferences over decisions, e.g. costs of resources and actions.
\end{enumerate}
	
User-centric explanations concern the system’s understanding of its user, and include aspects of cooperation and adaptation. 
At a high level, they aim to answer the following question:
\begin{itemize}
    \item Given a representation of the problem and of the user, e.g.\ their preferences or mental model, and given an object of interest (e.g.\ a query, a decision, a solution, a change, a desideratum), how are the user and the object related?
\end{itemize}

Explanations that deal with the (human) user's model or preferences are of particular interest in multi-agent environments. 
However, \citeauthor{Kraus.et.al:2020} stipulate in~\cite{Kraus.et.al:2020} that there has so far been little explainability in MAS. 
They claim explainability in MAS is more challenging than in other settings because “in addition to identifying the technical reasons that led to the decision, there is a need to convey the preferences of the agents that were involved.” 
\citeauthor{Murukannaiah.et.al:2020} echo this concern in~\cite{Murukannaiah.et.al:2020} from the points-of-view of multi-agent ethics, fairness etc. 
Several recent works suggest ways to address the challenges. 
At a high-level, in~\cite{Ciatto.et.al:2019} \citeauthor{Ciatto.et.al:2019} propose to integrate symbolic and connectionist approaches via a MAS to achieve explainability. 
\citeauthor{Kraus.et.al:2020}specifically propose to use ML for generating “personalized explanations that will maximize user satisfaction”, which necessitates collecting “data about human satisfaction from decision-making when various types of explanations are given in different contexts.”\cite[pp.~13534-13535]{Kraus.et.al:2020}
Along similar lines, in~\cite{Amir.et.al:2019} \citeauthor{Amir.et.al:2019} suggest research directions for agent strategy summarization to complement explainability in MAS. 
It remains to be seen what lines of research will be instigated by these recent calls to renew interest in MAS explainability and whether MR will play a significant role. 

Overall, the challenges with attaining user-centric explanations (and recognition of their current scarcity, at least in MR) have been brought forward in~\cite{Kraus.et.al:2020,Chakraborti.et.al:2020,Srinivasan:Chander:2020}. 
We hope that the developing landscape of MR explainability may soon allow (and require) an overview of the user-centric aspects of explanations.

\paragraph{Medium of explanations}

Relatedly, we did not consider the form of presenting explanations to the explainee, i.e.\ whether explanations are textual, graphical, etc.~\cite{Lacave:Diez:2002,Gonul:Onkal:Lawrence:2006,Sokol:Flach:2020,Mohseni:Zarei:Ragan:2020}. 
However important, these are largely human-AI interaction aspects, beyond the scope of this report.

\paragraph{Explanation desiderata}

In addition to such overarching properties as user-centrism and the categories of explanations suggested herein, 
explanations in AI can be studied and classified in terms of the \emph{desiderata} (or desirable properties) they fulfil. 
To appreciate the evolution of desiderata for AI explainability over the previous decades we invite the reader to consult the following works:~\cite{Johnson:Johnson:1993,Lacave:Diez:2002,Walton:2004,Lim:Dey:2010,Kulesza.et.al:2015,Lipton:2018,Hansen:Rieger:2019,Guidotti.et.al:2019,Sokol:Flach:2020}. 
We agree that a discussion of the various desiderata would be complementary to our overview, but it is beyond the scope of this work. 
We believe that our categorization works at a sufficiently high-level and is adequate for the purposes of this report.

\paragraph{Evaluation of explanations}
    
Relatedly, we did not consider aspects of explanation evaluation. 
We believe that, on the one hand, well-established computational metrics for systematic comparison and evaluation of explanations are generally lacking \cite{Hall.et.al:2019},\footnote{Though the literature on computational measures of explainability in ML is expanding, see e.g.~\cite{Carvalho:Pereira:Cardoso:2019,Mohseni:Zarei:Ragan:2020}.}
which we believe is especially true with respect to MR approaches to explainability. 
On the other hand, we believe that research on evaluation of, particularly, MR explainability approaches via human user studies is complicated and not yet mature either~\cite{Mohseni:Zarei:Ragan:2020}.

\paragraph{Scope of explanations}

We readily note that we have considered explanations of mostly \emph{local} scope, i.e.\ those that pertain to explaining the output $o$ of the AI system \AI\ given input $i$, as opposed to \emph{global} ones, i.e.\ those that pertain to explaining the AI system itself.
The former kind of explanations seem to prevail in MR, perhaps because it is common to consider MR techniques themselves to be more understandable than e.g.~ML ones. 
However, given the sophistication and complexity of modern MR techniques such as forms of \nameref{subsec:CP}, \nameref{subsec:Argumentation} or \nameref{subsec:Planning}, that proposition does not seem very compelling.

\paragraph{Applicability of categories}

We speculate that our loose categorization of explanations in AI applies to ML as well as to MR. 
For instance, post-hoc ML explanation techniques summarized in~\cite[p.~89, Figure 4]{Arrieta.et.al:2020}, are attributive, except for local explanations, which can also be seen as contrastive. 
It is reasonable to expect our categories to be applicable to other XAI approaches, given that our work builds on and borrows from recent XAI overviews. 
However, we by no means claim that our loose categorization is exhaustive: 
there may obviously be types of explanations that are not covered by our three families, e.g.\ explanations by example that are rather popular in ML~\cite{Mohseni:Zarei:Ragan:2020}. 
We leave it for future work to see how broadly our categorization applies to non-MR approaches.

\paragraph{Hierarchy of explanation families}

We see the three categories of explanations forming a hierarchy of increasing complexity in the following intuitive sense. 
Attributive explanations may act as components of contrastive explanations (e.g.\ attributions pro and con), and contrastive explanations can pave way for actionable explanations (e.g.\ contrasting outcomes lead to actions). 
This complexity also well reflects the maturity of different types of explanations, with attributive ones being the oldest and most pervasive, contrastive ones more recent and advanced, and actionable explanations arguably the most challenging and least explored. 
We have however not made these relationships more precise and leave it for future work.

\subsection{Terminology}
\label{subsec:Terminology}

There are a number of XAI terms carrying ambiguous meanings: explainability, interpretability, transparency, intelligibility, to name a few. 
Overall, they are somewhat intuitive and to an extent commonly understood, yet their precise meanings often diverge, arguably being dependent on the contexts of use. 
An in-depth discussion is beyond the scope here, but we state how we understand and differentiate the terms explainability and interpretability. 
To this end, we build on multiple XAI works referenced herein, but particularly follow~\cite{Hansen:Rieger:2019,Hall.et.al:2019,Rosenfeld:Richardson:2019,Arrieta.et.al:2020,Ciatto.et.al:2019,Ciatto.et.al:2020,Zhou:Danks:2020}. 
We specifically borrow from the ``Abstract Framework for Agent-Based Explanations in AI'' by~\citeauthor{Ciatto.et.al:2020} to differentiate between interpretability and explainability thus~\cite[p.1816-1817]{Ciatto.et.al:2020}. 

\begin{itemize}
    \item \emph{Interpretability} of an object $X$ pertains to an intelligent agent $A$ assigning a subjective meaning to $X$. 
    \item \emph{Explainability} of an object $X$ pertains to production of details and/or reasons of the functioning of $X$ so that an intelligent agent $A$ obtains (is either presented with, or can produce) a more interpretable object $X'$. 
\end{itemize}

Note that the act of interpreting (and understanding) an object requires an active entity $A$~\cite{Rosenfeld:Richardson:2019,Hall.et.al:2019,Ciatto.et.al:2020}. 
As such, interpretability of $X$ can be viewed as a “passive characteristic”~\cite{Arrieta.et.al:2020} of $X$ which enables (and is necessary for) understanding $X$. 
Thus, an AI system \AI\ is \emph{interpretable} to $A$ if $A$ can assign meaning to the components and internal representations of \AI. 
For example, let \AI\ have an internal representation/model $\M$. Then $A$ interprets (parts of) \AI\ by assigning meaning to $\M$, 
i.e.\ $A: \M \mapsto \M'$ where $\M'$ is $A$’s internal “mental model/alternative representation” of $\M$ (and thus \AI). 
Note that $\M'$ may not be formally describable, e.g.\ in case $A$ is human. 
We agree with~\cite{Ciatto.et.al:2020} that interpretability comes in degrees, specifically in allowing $A$ to deem distinct objects more or less interpretable. 
We however did not consider this aspect of interpretability in this report and thus do not delve in more details here. 

Now, explainability of an object $X$ can instead be viewed as an “active characteristic”~\cite{Arrieta.et.al:2020} by which $X$ makes itself interpretable (and consequently understandable) to entity $A$. 
Thus, an AI system \AI\ is \emph{explainable} to $A$ if \AI\ produces details and/or reasons of its functioning and outputs thereof that $A$ can assign meaning to. 
For example, \AI\ may have an “explainer component” $E$ which produces details and reasons $R$ of the functioning and outcomes of \AI. 
Briefly, $E: \AI \mapsto R$ so that $A$ can interpret $R$, i.e.\ $A: R \mapsto R'$ where $R'$ is $A$'s internal “understanding/representation” of the functioning and outcomes of $\AI$ (as supplied by $E$). 
Explainability is thus a more general concept and can be seen to cover~\cite{Hansen:Rieger:2019} or enable~\cite{Hall.et.al:2019,Rosenfeld:Richardson:2019}  interpretability.\footnote{To quote~\cite[p.~41-42]{Hansen:Rieger:2019}: 
    “The terms interpretability and explainability are often used interchangeably in the literature. However, $<\ldots>$ [t]he more general concept is explainability which covers interpretability, i.e., to communicate machine learning function to user, and completeness, i.e., that the explanation is a close enough approximation that it can be audited. The distinction is described: ‘...interpretability alone is insufficient. In order for humans to trust black-box methods, we need explainability -- models that are able to summarize the reasons for neural network behavior, gain the trust of users, or produce insights about the causes of their decisions. While interpretability is a substantial first step, these mechanisms need to also be complete, with the capacity to defend their actions, provide relevant responses to questions, and be audited’ [reference therein to~\cite{Gilpin.et.al:2018}]."
    For diverging opinions, note that e.g.~\cite{Miller:2019} equates explainability and interpretability.} 
Note, however, that the above are not formal notions and we use them to convey our intuitive understanding of the ambiguous concepts in question.\footnote{We also remark on the term transparency which seems to have an even more opaque (pun intended) meaning throughout the reviewed literature. 
    For instance, transparency can be a low-level property in opposition to opacity~\cite{Preece:2018} where “users can see some aspects of the inner state or functionality of the AI system”~\cite{Wang.et.al:2019}, referring to an AI system being interpretable~\cite{Guidotti.et.al:2019,Arrieta.et.al:2020} or requiring the connection between an algorithm and its interpretation to be both explicit and faithful~\cite{Rosenfeld:Richardson:2019}. 
    Or transparency can be a high-level concept that encompasses both interpretability and explainability via provision of information about the AI systems decision-making processes~\cite{Mohseni:Zarei:Ragan:2020} or amounts to “explaining the inner-workings of the systems to the user and informing about the intents of the agent”~\cite{Anjomshoae.et.al:2019}. 
    In view of this, we maintain that transparency is akin to interpretability, but that it does not require an active entity $A$ to interpret an object $X$, instead merely stipulating that $X$ inherently has the characteristic of enabling some $A$ to interpret $X$. 
    So, for instance, decision trees and logic programs are transparent, but neural networks (of non-trivial size) are not~\cite{Adadi:Berrada:2018}. 
    Nonetheless, interpreting and explaining decision trees or logic programs is non-trivial, and neural networks can still be explainable if details and reasons behind their functioning and outputs are produced and thereafter interpreted.}

\section{Conclusions}
\label{sec:Conclusions}

In this report, we have provided a high-level conceptual overview of selected Machine Reasoning (MR) approaches to explainability.
We have summarised what we believe are the most relevant MR contributions to Explainable AI (XAI), from early to modern MR research, perhaps with a stronger focus on the more recent studies. 
We have discussed explainability in MR branches of \nameref{subsec:Inference}, \nameref{subsec:LP}, \nameref{subsec:CP}, \nameref{subsec:ATP}, \nameref{subsec:Argumentation}, \nameref{subsec:DT} and \nameref{subsec:Planning} as well as the related areas of \nameref{subsec:Causal} and \nameref{subsec:RL}. 
In particular, we have seen that MR explainability approaches are suited not only for explainable MR (i.e.\ explaining MR-based AI systems) 
but also for explainability in other fields or areas of AI, such as Machine Learning (ML). 

We have loosely categorized the various kinds of explanations provided by MR approaches into three families of explanations: attributive, contrastive and actionable. 
Attributive explanations give details as to why an AI system yields a particular output given a particular input in terms of attribution and association of the (parts of the) system and the input with the output. 
This type of explanations have been studied since the very early MR and continue to be relevant and widely used in modern MR as well as its approaches to XAI at large. 
Contrastive explanations give details as to why an AI system yields one but not another output given some input in terms of reasons for and against different outputs. 
This type of explanations has been advocated for in MR for a long time too and appears in modern MR in the form of counterexamples, criticisms, counterfactuals and dialogues. 
Such and similar forms of contrastive explanations are being actively explored in MR and its applications to XAI. 
Finally, actionable explanations give details as to what can be done in order for an AI system to yield a particular output given input in terms of actions available to the system's user (human or AI, indifferently). 
This type of explanations should enable interventions to the system and eventually an interactive collaboration between the user and the system so as to reach desirable outcomes. 
We see actionable explanations as belonging to the frontier of MR explainability where novel research approaches and directions are being proposed. 

Our categorization of explanations is informed by the different types of questions about the AI system's workings which explanations seek to answer.
We have indicated some of the questions addressed in the overviewed MR branches and summarized the higher-level questions. 
In answering the latter, an explanation provides some details and reasons about the AI system's functioning and outputs. 
This addresses what we believe are the main purposes of explanations in XAI, namely to enable the user of an AI system to both understand the system and to do something with the explanation. 

The main lesson in writing this report was perhaps the (re)discovery of the evolution of XAI challenges and the wealth of MR approaches aiming to address those challenges. 
Importantly, we want to stress that XAI research in MR is very much active to date, as seen from our overview of modern MR explainability studies. 
Still, despite the advances over the years, challenges in MR explainability abound and it seems that lessons from the past hold well today too~\cite[p.~161]{Johnson:Johnson:1993}:
\begin{quote}
    If explanation provision is to become a characteristic feature of many future interfaces, then there is a special responsibility for researchers in both HCI [human-computer interaction] and AI to provide input to the debate about the nature of the explanations to be provided in future information systems. 
    The onus on us as researchers in the area is to ensure that we profit by past research on explanation provision, identify the strengths and weaknesses in present research and build on the strengths and address the problems in the future.
\end{quote}
We hope that this report will inform the XAI community about the progress in MR explainability and its overlap with other areas of AI, thus contributing to the bigger picture of XAI.

\newpage
\singlespacing
\small{
	\bibliographystyle{plainnat}
	\bibliography{references}
}

\end{document}